%% file: main.tex
\documentclass[journal]{IEEEtran}
\usepackage{amsmath,amsfonts}
\usepackage{algorithmic}
\usepackage{array}
\usepackage{textcomp}
\usepackage{stfloats}
\usepackage{url}
\usepackage{verbatim}
\usepackage{graphicx}

\usepackage{hyperref}
\usepackage{url}
\usepackage[ruled,vlined]{algorithm2e}
\usepackage{booktabs}
\usepackage{colortbl}
\usepackage{xcolor}
\usepackage{multirow}
\usepackage{amssymb} 
\usepackage{pifont} 
\usepackage[caption=false]{subfig}

\usepackage{bbm}
\usepackage{nicefrac}
\usepackage{overpic} 
\usepackage{makecell}
\usepackage{float}
\usepackage{cite}
\usepackage{threeparttable}

\usepackage{nomencl}
\makenomenclature

\def\Figref#1{Fig.~\ref{#1}}
\def\Tabref#1{Table~\ref{#1}}
\def\Secref#1{Section~\ref{#1}}
\def\Eqref#1{(\ref{#1})}

\definecolor{ourmethod}{gray}{0.93}
\definecolor{priormethod}{gray}{1} 

\definecolor{mydarkblue}{rgb}{0,0.08,0.45}
\definecolor{myredcolor}{RGB}{215,48,39}
\definecolor{mygreencolor}{RGB}{26,152,80}

\definecolor{DarkSeaGreen}{RGB}{143,188,143}
\definecolor{DarkGrey}{RGB}{169,169,169}
\definecolor{LightSkyBlue3}{RGB}{141,182,205}

\definecolor{orange}{RGB}{255,127,0}
\definecolor{DeepSkyBlue}{RGB}{0,191,255}
\definecolor{Purple}{RGB}{160,32,240}
\definecolor{LimeGreen}{RGB}{50, 205, 50}
\definecolor{Gold}{RGB}{255,215,0}
\definecolor{Silver}{RGB}{192,192,192}
\definecolor{Copper}{RGB}{185,123,88}

\urldef{\web}\url{https://zhangdongkun98.github.io/social-adv-flow}

\begin{document}
\title{Zero-shot Transfer Learning of Driving Policy via\\ Socially Adversarial Traffic Flow}
\author{
Dongkun Zhang,
Jintao Xue,
Yuxiang Cui,
Yunkai Wang,
Eryun Liu,
Wei Jing,
Junbo Chen, \\
Rong Xiong,
and Yue Wang

\thanks{
Dongkun Zhang, Yuxiang Cui, Yunkai Wang, Rong Xiong and Yue Wang are with the State Key Laboratory of Industrial Control and Technology, Zhejiang University, Hangzhou 310027, China.

Dongkun Zhang, Wei Jing and Junbo Chen are with the Department of Autonomous Driving Lab, Alibaba DAMO Academy, Hangzhou 310000, China.

Jintao Xue and Eryun Liu are with the College of Information Science and Electronic Engineering, Zhejiang University, Hangzhou 310027, China.

\textit{Corresponding author: Yue Wang.} (e-mail: wangyue@iipc.zju.edu.cn)

Video results, code, and simulator are available at \web.
}
}

\maketitle

\begin{abstract}
Acquiring driving policies that can transfer to unseen environments is challenging when driving in dense traffic flows.
The design of traffic flow is essential and previous studies are unable to balance interaction and safety-criticism.
To tackle this problem, we propose a socially adversarial traffic flow.
We propose a Contextual Partially-Observable Stochastic Game to model traffic flow and assign Social Value Orientation (SVO) as context. We then adopt a two-stage framework. In Stage 1, each agent in our socially-aware traffic flow is driven by a hierarchical policy where upper-level policy communicates genuine SVOs of all agents, which the lower-level policy takes as input. In Stage 2, each agent in the socially adversarial traffic flow is driven by the hierarchical policy where upper-level communicates mistaken SVOs, taken by the lower-level policy trained in Stage 1. Driving policy is adversarially trained through a zero-sum game formulation with upper-level policies, resulting in a policy with enhanced zero-shot transfer capability to unseen traffic flows.
Comprehensive experiments on cross-validation verify the superior zero-shot transfer performance of our method.

\end{abstract}


\input{sections/1-intro.tex}

\input{sections/2-related-work.tex}

\input{sections/3-method.tex}

\input{sections/4-results.tex}

\input{sections/5-conclusion.tex}

\bibliographystyle{IEEEtran}
\bibliography{reference}

\end{document}

%% file: sections/1-intro.tex





\nomenclature[1]{$G$}{Partially-Observable Stochastic Game.}
\nomenclature[1]{$G^{\boldsymbol{c}}$}{Contextual Partially-Observable Stochastic Game.}

\nomenclature[21]{$\mathcal{I}_n$}{Set of agent indices.}
\nomenclature[22]{$1$}{Index of ego agent (in Stage 2).}
\nomenclature[23]{$\mathcal{I}_n  \backslash \{1\}$}{Set of background agent indices.}

\nomenclature[3]{$\mathcal{S}$}{State space.}
\nomenclature[3]{$s$}{State.}

\nomenclature[41]{$\boldsymbol{\mathcal{A}}$}{Joint action space.}
\nomenclature[42]{$\mathcal{A}_i$}{Action space of agent $i$.}
\nomenclature[43]{$\boldsymbol{a}$}{Joint action.}
\nomenclature[44]{$a_i$}{Action of agent $i$.}

\nomenclature[51]{$\boldsymbol{\mathcal{O}}$}{Joint observation space.}
\nomenclature[52]{$\mathcal{O}_i$}{Observation space of agent $i$.}
\nomenclature[53]{$\boldsymbol{o}$}{Joint observation.}
\nomenclature[54]{$o_i$}{Observation of agent $i$.}

\nomenclature[61]{$\boldsymbol{R}$}{Set of agent-specific reward functions.}
\nomenclature[62]{$R_i$}{Reward function of agent $i$.}

\nomenclature[71]{$\boldsymbol{\mathcal{C}}$}{Joint context space.}
\nomenclature[72]{$\mathcal{C}_i$}{Context space of agent $i$.}
\nomenclature[73]{$\boldsymbol{c}$}{Joint context.}
\nomenclature[74]{$c_i$}{Context of agent $i$.}
\nomenclature[75]{$\boldsymbol{\bar{c}}$}{Genuine joint context.}
\nomenclature[76]{$\boldsymbol{\tilde{c}}$}{Mistaken joint context.}

\nomenclature[81]{$\boldsymbol{\beta}$}{Joint policy.}
\nomenclature[82]{$\beta_i$}{Policy of agent $i$.}
\nomenclature[83]{$\beta_i^u$}{Upper-level policy of agent $i$.}
\nomenclature[84]{$\beta_i^l$}{Lower-level policy of agent $i$.}
\nomenclature[85]{$\beta_i^*$}{Optimal policy of agent $i$.}
\nomenclature[86]{$\beta_i^{u, *}$}{Optimal upper-level policy of agent $i$.}
\nomenclature[87]{$\beta_i^{l, *}$}{Optimal lower-level policy of agent $i$.}

\printnomenclature[1.5cm]

\section{Introduction}

\IEEEPARstart{L}{ast} decade witnesses the success of reinforcement learning (RL) \cite{sutton2018reinforcement} in decision tasks such as Go \cite{silver2016mastering} and Atari \cite{mnih2013playing}, which inspires its trial in autonomous robots and vehicle \cite{ibarz2021train, kiran2021deep}. However, a barrier to deploying the learned driving policy is the dense traffic flows in real world, thus drawing the focus of the community \cite{bouton2019cooperation, saxena2020driving, bouton2020reinforcement, cai2022closing}. The difficulty is that when the well-trained driving policy is evaluated in traffic flows with unseen behavior patterns, the performance decreases significantly \cite{rajeswaranepopt, packer2018assessing, zhao2019investigating}. As we cannot assume the flow behavior in real world, it is essential to develop driving policy that is able to \textit{zero-shot transfer} among different traffic flows.

One solution to address this challenge is enriching the behavior patterns by introducing \textit{interaction} to the dense traffic flow, then learning the ego-driving policy from the flow. Prior works explore two methodologies to develop flow: analytical methods \cite{treiber2013traffic, cai2020summit} and multi-agent reinforcement learning (MARL) \cite{palanisamy2020multi, pal2020emergent, peng2021learning}. For analytical methods, Intelligent Driver Model (IDM) \cite{treiber2013traffic} uses one single differential equation to model agent-following behaviors and is used for RL training in \cite{wu2021flow, li2022metadrive}. However, as the number of agents increases, the stop-and-go nature \cite{stern2018dissipation} of IDM degenerates the whole flow efficiency, which we consider to be an indicator of insufficient interaction. To improve the traffic flow efficiency, \cite{cai2020summit} utilizes constrained geometric optimization in velocity space \cite{luo2022gamma} to capture the motion of traffic agents. For MARL methods, \cite{pal2020emergent} leverages a bi-level version of Proximal Policy Optimization (PPO) \cite{schulman2017proximal} for joint learning, which also improves the efficiency of the traffic flow. But these efficiency-only optimization methods cause the agents in these traffic flows to be homogeneous, thereby reducing the diversity of behavior, which still reflects insufficient interaction. \cite{peng2021learning} tackles this problem by incorporating Social Value Orientation (SVO) \cite{liebrand1984effect}, which describes how an agent values the individual and group benefit. By assigning different SVOs to multiple agents, diverse social behaviors are expected to emerge after MARL training. Unfortunately, for both lines of methods, little attention has been paid to zero-shot transfer evaluation of ego policy learned from their traffic flows.

In addition to the interaction in the flow, as \cite{Hanselmann2022king} mentions, the zero-shot transfer capability of ego-driving policy learned from the flow may remain low due to insufficient exposure to near-accident \textit{safety-critical} data. Following this idea, some works employ robust adversarial reinforcement learning (RARL) \cite{pinto2017robust, pan2019risk, vinitsky2020robust, oikarinen2021robust}. These methods model differences between training and testing environments as disturbances towards driving policy. To achieve this, existing works \cite{wachi2019failure, ren2020improving, chen2021adversarial, sharif2021evaluating, ijcai2022-430} leverage several background agents to attack the ego agent by causing its failure. Although working well in sparse traffic situations, this pipeline cannot scale up to dense traffic flows. Increasing the number of attacking agents makes it difficult for the ego agent to resist such strong disturbances, and also prevents multiple adversarial agents from reaching their individual goals, which actually breaks the interaction. As a result, the problem is raised as: \textit{Is it possible to build a dense traffic flow that is interactive and safety-critical simultaneously?}

\begin{figure*}[t]
\begin{center}
\includegraphics[width=1.0\textwidth]{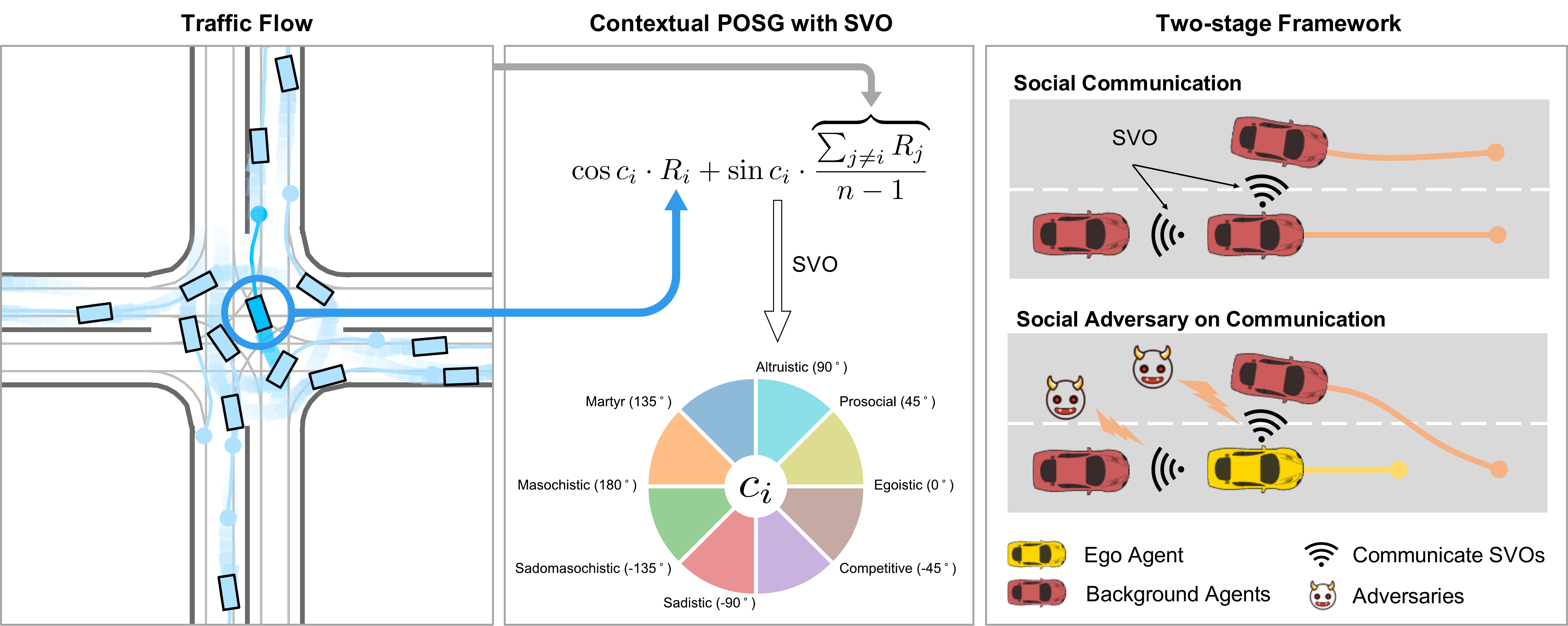}
\end{center}
\vspace{-0.3cm}
\caption{Overview of our method. \emph{Left}: Traffic flow at an unsignalized intersection. \emph{Center}: We propose a Contextual POSG to model traffic flow. The context is assigned with SVOs. \emph{Right}: We adopt a two-stage framework. In Stage 1, agents communicate their genuine SVOs. In Stage 2, adversaries are introduced to produce mistaken SVOs, disturbing the SVO communication.}
\label{figure:overview}
\end{figure*}

In this paper, we introduce the social-level communication and adversary to a dense traffic flow, in order to improve the zero-shot transfer of the learned ego-driving policy. The core idea is to develop social communication among agents and attack the social communication between the ego and background agents, with the aim of balancing the interaction and safety-criticism. As shown in \Figref{figure:overview}, we state the traffic flow (including ego agent) as a Contextual Partially-Observable Stochastic Game (CPOSG).
Each agent in CPOSG is driven by a hierarchical policy with the context as the intermediate variable, which we assign with all agents' SVOs. Thus each agent is enabled to communicate its SVO and behaves accordingly. Based on this model, we propose a two-stage method: In Stage 1, all agents in the traffic flow share their SVOs to achieve higher efficiency and diversity. In Stage 2, an adversary is injected into the genuineness of the communicated SVO between the ego agent and the other agents, formulating a zero-sum game. In this way, we expect the dense flow to be both interactive and safety-critical, emerging an ego-driving policy can gain higher robustness for zero-shot transfer. In the experiment, we show the efficiency and diversity of the dense traffic flow both quantitatively and qualitatively, validating the effectiveness of communication on the interaction. We then conduct the zero-shot transfer by massive cross-validation where we evaluate each trained ego-driving policy using $8000$ test episodes per scenario. By evaluating $6$ policies on $4$ scenarios, in total $192000$ episodes, we arrive at the results that the policy trained in our traffic flow shows statistically significant better performances in almost all unseen traffic flows, verifying the effectiveness of the proposed method. In summary, the contributions of this paper are as follows:
\begin{itemize}
    \item A novel formulation of traffic flow as a Contextual POSG (CPOSG) problem, with a hierarchical policy whose context is assigned with SVOs.
    \item An interactive traffic flow that can communicate genuine SVOs is proposed to improve the whole flow efficiency while keeping the diversity.
    \item An adversarial training method by communicating mistaken SVOs is proposed to generate the safety-critical data for training robust driving policy.
    \item Comprehensive evaluations verifying the performance of our traffic flow, and massive cross-validation verifying the zero-shot transfer capability of our method.
\end{itemize}

%% file: sections/2-related-work.tex
\def\citeauthor#1{\textit{et al.} \cite{#1}}

\section{Related Work}  \label{section-rw}

\subsection{Traffic Flows} \label{section-rw:traffic-flows}

Many works focus on mixed-autonomy traffic and use one or several autonomous vehicles (AVs) to regulate human-driven traffic flows (HDFs) via control-theoretic and RL frameworks.
Researchers in \cite{cui2017stabilizing, stern2018dissipation} analyze the potential of a single AV in stabilizing the whole HDF via control theory while several AVs are leveraged in \cite{talebpour2016influence, wu2018stabilizing, yao2019stability, zheng2020smoothing}.
Wu \citeauthor{wu2017emergent} utilizes Trust Region Policy Optimization (TRPO) \cite{schulman2015trust} to improve ring road efficiency. Vinitsky \citeauthor{vinitsky2018benchmarks} presents an RL-based benchmark for traffic control. Researchers in \cite{cui2021scalable, zhang2021learning} use MARL algorithms to control several AVs and mitigate traffic congestion caused by HDF. Importantly, Wu \citeauthor{wu2021flow} demonstrates that increasing the proportion of AVs is beneficial to the overall efficiency of traffic flow, Therefore, another line of work aims to improve the performance of traffic flows consisting of pure AVs. Cai \citeauthor{cai2020summit} use a road-aware crowd behavior model based on \cite{luo2022gamma} to generate interactive behaviors of traffic agents. Pal \citeauthor{pal2020emergent} develops a bi-level PPO and applies IPL to learn a traffic flow. Peng \citeauthor{peng2021learning} leverages SVO to build an efficient traffic flow where all agents share the same distribution of SVO.

\subsection{Social Value Orientation}

\begin{figure*}[ht]
    \begin{center}
    \includegraphics[width=1.0\textwidth]{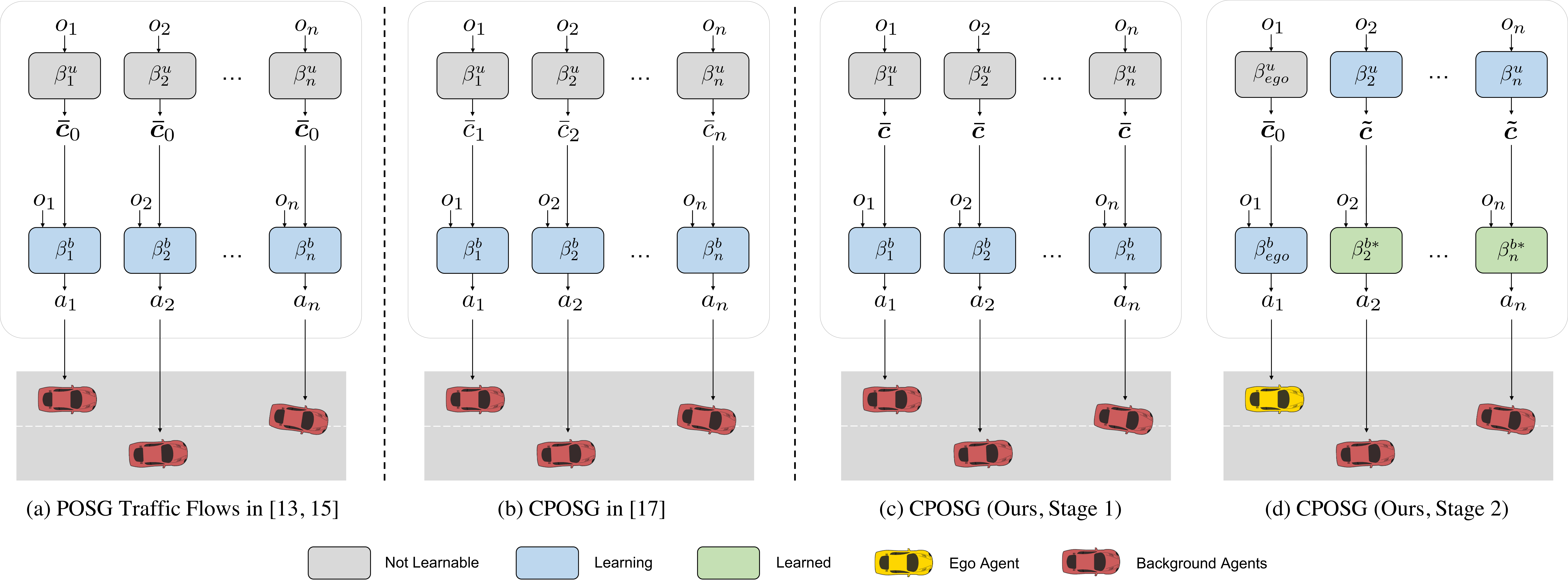}
    \end{center}
\vspace{-0.3cm}
\caption{Hierarchical policy under CPOSG. Each agent $i$ is driven by a hierarchical policy that takes observation $o_i$ as input and outputs action $a_i$. $\boldsymbol{c}$ is the context of CPOSG. $\boldsymbol{c}_0$ represents constant context while others represent variable contexts. A bar superscript represents genuine context and a tilde superscript represents mistaken context.}
\label{figure:policy}
\end{figure*}

SVO is originally a concept in social psychology that captures an agent's preference about how to balance rewards between the self and another agent \cite{liebrand1984effect, liebrand1988ring, mcclintock1989social}. 
Schwarting \citeauthor{schwarting2019social} introduces this concept to the field of autonomous driving and formulates a best-response game in merging lane and unprotected left turn.
Buckman \citeauthor{buckman2019sharing} leverages a central coordinator to reduce system blocks while ensuring each agent increases its own utility.
Researchers in \cite{toghi2021cooperative, toghi2022social, toghi2021altruistic} leverage MARL to improve overall performance in mixed traffic.
Zhao \citeauthor{zhao2021yield} proposes a parallel-game-based method to capture interactions among agents and estimate the SVO of interacting agent in unprotected left-turn. To represent different reasoning abilities, level-$k$ gaming is adopted in \cite{wang2022comprehensive}.
Crosato \citeauthor{crosato2022interaction} focuses on vehicle-pedestrian interaction and exhibits different driving styles by varying SVO.
Peng \citeauthor{peng2021learning} builds an efficient traffic flow where agents are learned independently, followed by using a central coordinator to vary SVOs, leading to better coordination. In this paper, we improve efficiency via communication, without the need for a central coordinator.

\subsection{Adversarial Training}

A common way to acquire robust policy is applying RARL \cite{pinto2017robust, pan2019risk, vinitsky2020robust, oikarinen2021robust}. RARL methods regard model errors between training and evaluating environments as extra disturbances and introduce an adversarial policy to exert disturbances on dynamics parameters such as joint force and friction coefficient. This formulates a zero-sum game between ego policy and adversarial policy. Adversarial policy aims to reduce the performance of the ego policy, while the ego policy learns to act robustly by resisting the disturbances introduced by the adversarial policy.

Researchers in autonomous driving also follow this pipeline \cite{wachi2019failure, ren2020improving, chen2021adversarial, sharif2021evaluating, ijcai2022-430}. Instead of disturbing dynamics parameters, most works leverage few background agents to attack ego agent.
Adversarial policies in \cite{ren2020improving, chen2021adversarial, sharif2021evaluating, ijcai2022-430} are optimized to collide with ego agent, while researchers in \cite{wachi2019failure, ijcai2022-430} attempt to expel ego agent from drivable areas. 
Formulating a zero-sum game leads to unreasonable behaviors of background agents. For instance, an adversarial agent may directly rush into the opposite lane and crash with ego agent, totally ignoring traffic rules which the ego agent obeys. Therefore, researchers in \cite{wachi2019failure, chen2021adversarial} relax the zero-sum assumption and encourage adversarial agents to obey traffic rules such as arriving at their own goals as fast as possible and driving in right lane in addition to attack ego agent.
However, we argue that using attacking agents to interfere with ego agent deliberately still provides strong disturbances, failing to improve the zero-shot transfer performance of ego-driving policy.

%% file: sections/3-method.tex
\section{Statement of Traffic Flow}  \label{section-ps}

\subsection{Traffic Flow}
\textbf{POSG Traffic Flow:} Agents in traffic flow are often partially observed since the intentions of other agents are unavailable and each agent makes decisions from its own perspective. Besides, instead of achieving a common goal for the whole system, each agent has individual goals like arriving at its own destination.
Therefore, a traffic flow system is typically formulated as Partially-Observable Stochastic Game (POSG) \cite{oliehoek2016concise}.
Formally, POSG is a tuple $G = \left<\mathcal{I}_n, \mathcal{S}, \boldsymbol{\mathcal{A}}, P, \boldsymbol{R}, \rho_0, \boldsymbol{\mathcal{O}}, \gamma, T\right>$. $n$ is the number of agents.
$\mathcal{I}_n = \{1,2,\dots,n\}$ denotes the set of agent indices. $\mathcal{S}$ is the state space and $s \in \mathcal{S}$ is the state of the whole system. $\boldsymbol{\mathcal{A}} = \times_{i \in \mathcal{I}_n} \mathcal{A}_i$ is the joint action space of $n$ agents and $\mathcal{A}_i$ is the action space available to agent $i$. $\boldsymbol{a} = \left< a_1, a_2, \dots, a_n \right> \in \boldsymbol{\mathcal{A}}$ is the joint action and $a_i \in \mathcal{A}_i$ is the action of agent $i$. 
$P(s' | s, \boldsymbol{a})$ is the state transition probability where $s', s \in \mathcal{S}, \boldsymbol{a} \in \boldsymbol{\mathcal{A}}$.
$\boldsymbol{R} = \{ R_1, R_2, \dots, R_{n} \} $ denotes the set of agent-specific reward functions and $R_i(s, \boldsymbol{a})$ is bounded for all $i \in \mathcal{I}_n$. Note that each agent $i$ receives an individual reward from its own reward function $r_i = R_i(s, \boldsymbol{a})$. $\rho_0(s)$ is the initial state distribution.
$\boldsymbol{\mathcal{O}} = \times_{i \in \mathcal{I}_n} \mathcal{O}_i$ is the joint observation space and $\mathcal{O}_i$ is the observation space available to agent $i$.
$\boldsymbol{o} = \left< o_1, o_2, \dots, o_n \right> \in \boldsymbol{\mathcal{O}}$ is the joint observation and $o_i \in \mathcal{O}_i$ is the observation of agent $i$. 
$\gamma \in (0, 1]$ is the discount factor and $T$ is the time horizon.
We define trajectory over one episode as $\tau = \left<\tau_0, \tau_1, \dots, \tau_T\right>$ where $\tau_t = \left<s_t, \boldsymbol{a}_t\right>$.
In $G$, each agent $i$ utilizes an observation-based policy $\beta_i (a_i | o_i)$ to maximize its own expected cumulative reward:
\begin{align}
    \beta_i^* = \mathop{\arg\max}_{\beta_i} \mathbb{E}_{\tau \sim \boldsymbol{\beta}} [\sum_{t=0}^{T}\gamma^{t} R_i], \quad i \in \mathcal{I}_n   \label{equation:object-posg}
\end{align}
where $\beta_i^*$ is the optimal policy for agent $i$, $\boldsymbol{\beta} = \{ \beta_1, \beta_2, \dots, \beta_{n} \} $ is the joint policy.
$\beta_i$ is typically represented by a neural network.

This formulation can cover most prior works on building dense traffic flows. 
For instance, traffic flow with IDM policy \cite{treiber2013traffic} can be regarded as a special case of POSG where optimal policy is defined as a differential equation and reward function is not given explicitly. \cite{palanisamy2020multi} uses POSG and applies several MARL settings to obtain traffic flow. However, real-world traffic flows are more diverse, making it difficult for a single POSG to capture. Therefore, a more general formulation is needed.

\textbf{Contextual POSG Traffic Flow:} To improve diversity, we introduce context into POSG, namely Contextual POSG (CPOSG) as $G^{\boldsymbol{c}} = \left<\boldsymbol{\mathcal{C}}, \mathcal{M}, \mathcal{I}_n, \mathcal{S}, \mathcal{A}, \mathcal{O}, \gamma, T\right>$. $\boldsymbol{\mathcal{C}} = \times_{i \in \mathcal{I}_n} \mathcal{C}_i$ is the joint context space and $\mathcal{C}_i$ is the agent-wise context space of agent $i$.
$\boldsymbol{c} = \left< c_1, c_2, \dots, c_n \right> \in \boldsymbol{\mathcal{C}}$ is the joint context and $c_i \in \mathcal{C}_i$ is the context of agent $i$. Context $\boldsymbol{c}$ remains constant throughout an episode.
Similar to the definition of Contextual MDP \cite{hallak2015contextual}, $\mathcal{M}$ is a function that tells how a context $\boldsymbol{c}$ modulates POSG components, i.e., $\mathcal{M}(\boldsymbol{c}) = \left<P^{\boldsymbol{c}}, \boldsymbol{R}^{\boldsymbol{c}}, \rho_0^{\boldsymbol{c}}\right>$ where $\boldsymbol{R}^{\boldsymbol{c}} = \{ R_1^{\boldsymbol{c}}, R_2^{\boldsymbol{c}}, \dots, R_{n}^{\boldsymbol{c}} \} $. The objective for each agent $i$ in $G^{\boldsymbol{c}}$ is as follows:
\begin{align}
    \beta_i^* = \mathop{\arg\max}_{\beta_i} \mathbb{E}_{\tau \sim \boldsymbol{\beta}} [\sum_{t=0}^{T}\gamma^{t} R_i^{\boldsymbol{c}}], \quad i \in \mathcal{I}_n, \boldsymbol{c} \in \boldsymbol{\mathcal{C}}   \label{equation:object-cposg}
\end{align}

As one can see, context $\boldsymbol{c}$ allows for diversity of multi-agents in the traffic flow by applying $\mathcal{M}$. Through proper design of joint context space, CPOSG $G^{\boldsymbol{c}}$ can cover the origin POSG $G$ with a context, say $\boldsymbol{\bar{c}}_0$, i.e., $G = G^{\boldsymbol{c}} | _{\boldsymbol{c} = \boldsymbol{\bar{c}}_0}$. 
To improve the diversity of traffic flow, we can uniformly sample joint context space $\boldsymbol{\mathcal{C}}$ to obtain $\boldsymbol{c}$ before each episode.

\subsection{Hierarchical Policy}
Under CPOSG, observation-based policy $\beta_i (a_i | o_i)$ reveals a hierarchical architecture by introducing context $\boldsymbol{c}$ as intermediate variable:
\begin{equation}
    \beta_i (a_i | o_i) = \int_{\boldsymbol{c} \in \boldsymbol{\mathcal{C}}} \underbrace{\beta^l_i(a_i | o_i, \boldsymbol{c})}_{\text{lower-level}} \underbrace{\beta^u_i(\boldsymbol{c} | o_i)}_{\text{upper-level}} d\boldsymbol{c} \label{equation:hierarchical}
\end{equation}
where upper-level policy is conditioned on the observation of agent $i$ and produces joint context, which is passed into the lower-level policy to produce the action of agent $i$.

\textbf{Constant vs. Variable Context:}
Since POSG can be viewed as a CPOSG with a \textit{constant} context $\boldsymbol{\bar{c}}_0$, its policy is a special case of \Eqref{equation:hierarchical} as shown in block (a) of \Figref{figure:policy}.
The policy proposed in \cite{peng2021learning} can also be regarded as a special case of \Eqref{equation:hierarchical} as shown in block (b) of \Figref{figure:policy}.
Specifically, their upper-level policy controls the visibility of context $\boldsymbol{c}$ where only genuine self context $\bar{c}_i$ is visible to agent $i$, while others not, i.e., $\beta^u_i(\boldsymbol{c} | o_i) = \beta^u_i(\boldsymbol{c}) \propto \delta({c}_i - \bar{c}_i)$, where $\delta$ is the Dirac delta function. Through this definition, the observation-based policy is reduced to its lower-level policy which is dependent on its own \textit{variable} context $\bar{c}_i$, i.e., $\beta_i(a_i | o_i) = \beta^l_i(a_i | o_i, \bar{c}_i)$. By uniformly sampling context space at the start of each episode, this policy formulation improves diversity.

\textbf{Invisible vs. Visible Communication:}
Based on the above discussion, the general upper-level policy can be regarded as a communication policy that receives all agents' contexts and selectively sends them to the lower-level policy. From this viewpoint, \cite{peng2021learning} uses a self-visible communication policy that only makes its own context visible. This approach leads to limited efficiency as mentioned in their work and we attribute this limitation to the low degree of visibility. To improve efficiency, we introduce a \textit{fully-visible} communication policy where all agents' genuine contexts are visible to its lower-level policy, i.e., $\beta^u_i(\boldsymbol{c} | o_i) = \beta^u_i(\boldsymbol{c}) = \delta(\boldsymbol{c} - \boldsymbol{\bar{c}})$. The lower-level policy $\beta^l_i (a_i | o_i, \boldsymbol{\bar{c}})$ is learned via \Eqref{equation:object-cposg} and efficiency of the whole traffic flow is improved while maintaining diversity. The hierarchical policy is shown in block (c) of \Figref{figure:policy}. This step is named Stage 1, which is introduced in \Secref{section-method-flow}.

\textbf{Genuine vs. Mistaken Communicated Context:}
When the communication policy above passes \textit{genuine} contexts, agents in traffic flow coordinate with each other and the whole system is efficient. On the contrary, communication policy can also pass \textit{mistaken} contexts to its lower-level policy, which offers an opportunity to inject adversary into our traffic flow. To achieve this, we introduce a learnable communication policy and train it in an adversarial manner where a zero-sum game between ego-driving policy and upper-level policies of other agents is formulated.
By freezing lower-level policies learned in Stage 1, the effect of the mistaken contexts produced by adversarial communication policy is constrained and avoids over-strong adversarial attacks in the dense traffic flow. The hierarchical policy is shown in block (d) of \Figref{figure:policy}. This Stage 2 is introduced in \Secref{section-method-adv}.

\section{Stage 1: Socially-aware Traffic Flow} \label{section-method-flow}

\subsection{System Modeling} \label{section-method-flow:comp}

\textbf{State Space:}
At a given time step, the state $s \in \mathcal{S}$ contains the set of static elements and dynamic elements. 
Static elements include lane centerlines, sidelines, and global paths of all agents. Dynamic elements include current poses and velocities of all agents.
Centerline consists of a sequence of polylines $\left\{\mathcal{P}_1, \mathcal{P}_2, \dots, \mathcal{P}_{J}\right\}$. Each polyline $\mathcal{P}_j$ consists of a sequence of vectors $\left\{\boldsymbol{v}_1, \boldsymbol{v}_2, \dots, \boldsymbol{v}_{K_{j}}\right\}$. A static vector $\boldsymbol{v}_k$ is defined as $\left[ x_k, y_k, \theta_k, l_j, k \right]$ where $(x_k, y_k)$ is the location, $\theta_k$ is the orientation, and $l_j$ is the lane width of polyline $j$. The definition of sideline is similar, except that the lane width of each polyline in sideline is $0$.
The global path of each agent $i$ is unchanged during one episode and also consists of a sequence of vectors.
The dynamic element for each agent $i$ is a vector $\boldsymbol{v}_i = \left[ x_i, y_i, \theta_i, v_i \right]$ where $v_i$ represents the speed of agent $i$.

\textbf{Observation Space:}
At time step $t$, observation $o_i \in \mathcal{O}_i $ can only observe system state $s$ partially. On the one hand, global paths of other agents are not available for agent $i$. On the other hand, all vectors are transformed into its own coordinate system w.r.t. current pose $(x_i, y_i, \theta_i)$. Furthermore, we add historical poses and velocities into the observation, in order to capture high-order dynamic information of agents. In this way, information of each agent $j \in \mathcal{I}_n$ is represented by a polyline $\mathcal{P}_j = \left\{\boldsymbol{v}_1, \boldsymbol{v}_2, \dots, \boldsymbol{v}_{H}\right\}$ where $H$ is the historical horizon. A vector $\boldsymbol{v}_h$ is defined as $\left[ x_h, y_h, \theta_h, v_h, h \right]$ where $h$ represents vector at time $t - h + 1$.

\textbf{Action Space:}
The action $a_i \in \mathcal{A}_i$ is $(v_{i}^r, \sigma_{i})$ where $v_{i}^r \in [0, v_{max}]$ is reference speed and $\sigma_i \in [-\sigma_{max}, \sigma_{max}]$ is steer. We introduce a PID controller that outputs acceleration and controls bicycle model \cite{kong2015kinematic} to track given reference speed. Input of bicycle model is acceleration and steer. State is pose and speed.

\subsection{Context-Modulated Reward} \label{section-method-flow:comp-reward}

\begin{figure*}[ht]
    \begin{center}
    \includegraphics[width=1.0\textwidth]{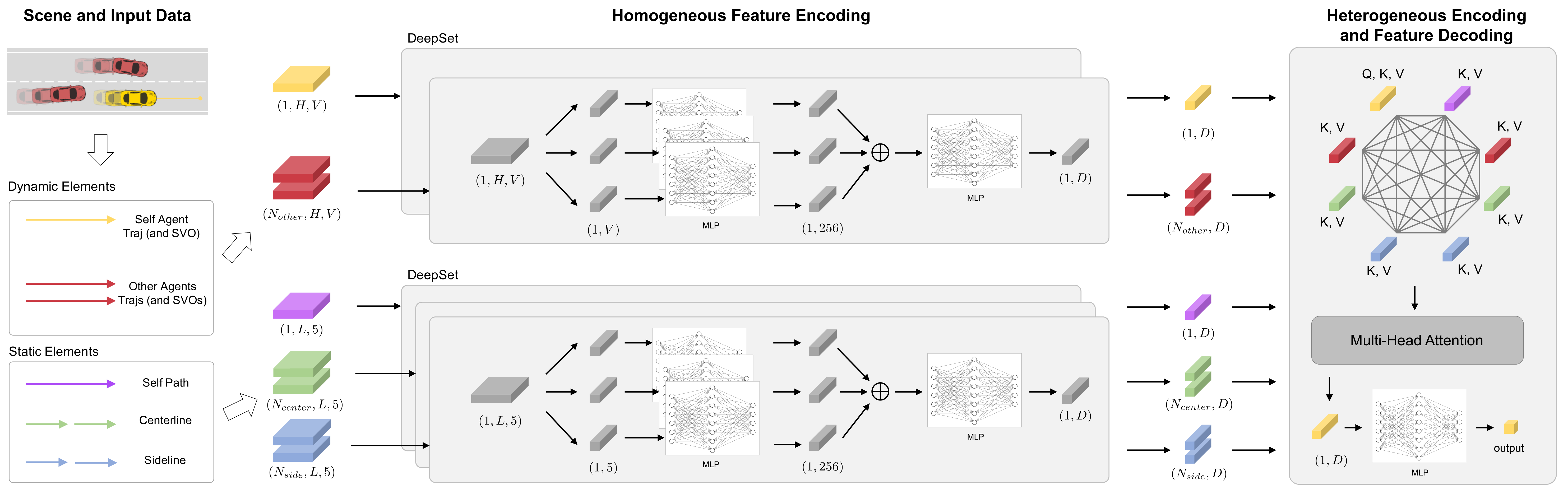}
    \end{center}
\vspace{-0.3cm}
\caption{Policy architecture for upper-level and lower-level policy. Upper-level policy takes observation as input and output SVOs. Lower-level policy takes observation and SVOs as input and output action. The architecture consists of feature encoding and feature decoding.}
\label{figure:net_fe}
\end{figure*}

In our socially-aware traffic flow, each agent takes into account not only its own reward but also the rewards of other agents by utilizing contextual information. The efficiency of traffic flow is improved by designing upper-level policy as communication policy. In this part, we define context $c_i$ as social preference which composes individual reward $R_i$ into socially-composed reward $R_i^{c_i}$. For each agent, its lower-level policy maximizes cumulative socially-composed reward.

\textbf{Individual Reward:}
The individual goal of each agent in dense traffic flow is homogeneous. For instance, all agents expect to successfully finish the task as fast as possible. Therefore, all agents share the same reward structure.
For each agent $i$, we design a near-sparse reward function containing a dense reward function $R_{i,speed}$ for incentive driving fast and a sparse reward function $R_{i,fail}$ for penalizing catastrophic failures. Catastrophic failures include collision with other agents, deviation from drivable area, driving too far from global path, and crashing into wrong lane. If agent $i$ encounters catastrophic failures, it is terminated and removed from the environment. Episode ends when all agents are terminated. The detailed definition of termination is introduced in \Secref{section-exp:sim}.

The dense reward function $R_{i,speed}$ is defined as follow:
\begin{align}
    R_{i,speed} = 2 \frac{v_i}{v_{max}} -1, \quad v_i \in [0, v_{max}]
\end{align}
where $v_i$ and $v_{max}$ is the agent's current and maximum speed respectively. Since $v_i \in [0, v_{max}]$, $R_{i,speed} \in [-1,1]$ is bounded. Intuitively, this reward term encourages high driving speed and penalizes low driving speed.

The sparse reward function $R_{i,fail}$ is defined as follow:
\begin{equation}
\begin{split}
    R_{i,fail} = - \mathbf{1}(
    \text{Collision}(i) \ \lor \ 
    \text{Off Road}(i) \ \lor \  \\
    \text{Off Route}(i) \ \lor \ 
    \text{Wrong Lane}(i))
\end{split}
\end{equation}
where $\mathbf{1}$ is the indicator function. If any catastrophic failure happens, $-1$ penalty is assigned to agent $i$.

The overall reward function is the combination of above items:
\begin{equation}
    R_i = \omega_1 R_{i,speed} + \omega_2 R_{i,fail}   \label{equation:reward}
\end{equation}
Note that $R_{i,speed}$ is accumulated each time step while $R_{i,fail}$ is received only at the end of each episode, we set $\omega_2 \gg \omega_1$ in order to avoid the circumstances that the agent sacrifices safety to drive fast.

\textbf{Socially-Composed Reward:}
We introduce SVO as the agent-special context $c_i$. SVO is a scalar ranging from $-180^{\circ}$ to $180^{\circ}$ and is used to capture social preferences as a combination of two terms including the ego agent's reward and another agent's reward, as shown in the center bottom of \Figref{figure:overview}. Since agents do not harm each other intentionally in our traffic flow, we define $c_i \in \mathcal{C}_i = [0^\circ, 90^\circ]$. SVO only acts on reward function $\boldsymbol{R}$, not on state transition probability $P$ and initial state distribution $\rho_0$. Furthermore, the reward function of agent $i$ is only affected by its specific context $c_i$. Accordingly, each agent socially composes multi-agent individual rewards as below:
\begin{equation}
    R_i^{\boldsymbol{c}} = R_i^{c_i} = \cos c_i \cdot R_i + \sin c_i \cdot \underbrace{\frac{ \sum_{j \neq i} R_j }{n-1}}_{R_{\neg i}}  \label{equation:reward-svo}
\end{equation}
where $R_{\neg i}$ is the average reward function of all agents except agent $i$. \Eqref{equation:reward-svo} forces agent $i$ to consider other agents' individual rewards, indicating that if another agent fails, agent $i$ should bear some responsibility, whereas driving fast also owes to agent $i$ to some extent.
From this definition, the POSG traffic flow is a CPOSG flow with zero context.

In this stage, communication policy is not learnable as shown in block (c) of \Figref{figure:policy}. The observation-based policy reduces to lower-level policy which maximizes cumulative socially-composed reward, obtaining the socially-aware policy $\beta_i^{l,*}$:
\begin{align}
    \beta_i^{l,*} = \mathop{\arg\max}_{\beta_i^l} \mathbb{E}_{\tau \sim \boldsymbol{\beta}} [\sum_{t=0}^{T}\gamma^{t} R_i^{c_i}], \quad i \in \mathcal{I}_n, c_i \in \mathcal{C}_i   \label{equation:object-socially-aware}
\end{align}

\subsection{Network Architecture of Lower-level Policy}  \label{section-method-flow:nn}

The neural network of lower-level policy $\beta^l_i(a_i | o_i, \boldsymbol{\bar{c}})$ consists of two modules: feature encoding and feature decoding.
Feature encoding module aggregates information from observation and SVOs to produce a latent representation with fixed size. Feature decoding module maps the latent representation to the action of agent $i$.

For feature encoding, we utilize a hierarchical framework as shown in \Figref{figure:net_fe}. We use two DeepSets \cite{zaheer2017deep} to encode homogeneous information within dynamic and static elements separately, followed by Multi-Head Attention (MHA) \cite{vaswani2017attention} to further encode heterogeneous information among different elements. Specifically, each element is a 3-dimensional tensor. The first dimension is the number of polylines. The second dimension is the length of each polyline. The third dimension is the length of each vector in one polyline. As described in \Secref{section-method-flow:comp}, the length of static vectors is $5$. The length of dynamic vectors $V$ is $6$ for lower-level policy where the last scalar is SVO, concatenated with dynamic vector $\boldsymbol{v}_h$.
Homogeneously, DeepSet encodes features within each polyline and outputs a fixed-size $D$-dimension feature. Each vector in one polyline is passed through a Multi-Layer Perception (MLP), followed by a $\text{sum}$ operator to fuse information from each vector. Another MLP is introduced to add nonlinearity, resulting in the $D$-dimension feature. After homogeneous encoding, information on each polyline is represented as a fixed-size vector. The relation among polylines is modeled as a fully-connected graph and we use MHA to encode features from this graph. Feature from the agent itself is the query and all features act as keys and values. For feature decoding, we use a simple $3$-layer MLP to output the action for agent control.

\subsection{Training}

We apply Independent Policy Learning (IPL) \cite{tan1993multi} to obtain the optimal lower-level policy. IPL decomposes the $n$-agent MARL problem into $n$ single-agent problems where other agents are treated as part of environment from the perspective of each agent. We utilize Soft-Actor-Critic (SAC) \cite{haarnoja2018soft} to train each agent. We first collect data by rolling out the hierarchical policy on CPOSG for several episodes and then optimize the lower-level policy $\beta^l_{i}$ with collected data. This procedure is repeated and the trained lower-level policy is socially-aware policy $\beta^{l,*}_{i}$.

\textbf{Implementation Details:}
To guarantee the scalability of our traffic flow, we further apply parameter sharing where all agents maintain a single policy network, i.e., $\beta^l_i (a_i | o_i, \boldsymbol{\bar{c}}) = \beta^l(a_i | o_i, \boldsymbol{\bar{c}})$, data collected from all agents are stored in the same experience buffer and used to optimize the same policy network. Besides, as clarified in \Secref{section-method-flow:comp}, observation $o_i$ contains all the dynamic and static vectors in one scenario, which is fed into the neural network, making it time-and-space consuming to train and deploy lower-level policy. To simplify the training, we clip observation $o_i$ and only keep vectors within a certain distance relative to agent $i$. Each agent can only communicate contexts with its surrounding agents. This is reasonable since human drivers tend to focus on neighbor drivers and road structure while neglecting distant information.

\section{Stage 2: Socially Adversarial Policy Training} \label{section-method-adv}

\subsection{Injecting Adversary}  \label{section-method-adv:problem}

\textbf{Driving Policy Formulation:} In this Stage 2, we define agent $1$ as the ego agent and remaining agents with indices $i \in \mathcal{I}_n  \backslash \{1\}$ are background agents in traffic flow.
Background agents can be regarded as part of the state transition probability from the perspective of ego agent. Therefore, training an ego-driving policy is a classic single-agent task. Note that an ego-driving policy that can be deployed in real world cannot depend on the communication, we set arbitrary constant context $\bar{\boldsymbol{c}}_0$ to the upper-level policy of ego agent as $\beta^u_{1} = \delta(\boldsymbol{c} - \bar{\boldsymbol{c}}_0)$. The lower-level policy of ego agent is then $\beta^l_{1} = \beta^l_{1}(a_1 | o_1, \bar{\boldsymbol{c}}_0)$. As the lower-level policy is further learned in Stage 2, the arbitrary constant context actually makes no difference. The learned lower-level policy of ego agent is finally regarded as the driving policy for zero-shot transfer.

\textbf{Social Adversary:}
To exhibit adversarial behaviors towards ego agent, we build our socially adversarial traffic flow on top of socially-aware traffic flow in \Secref{section-method-flow}. In this stage, the lower-level policy of each background agent is the learned socially-aware policy $\beta^{l,*}_i$, while, the upper-level policy for each background agent is learnable and passes mistaken SVOs to its socially-aware policy according to its observation, i.e., $\beta_i^{u} = \beta_i^{u} (\boldsymbol{\tilde{c}} | o_i)$ where $\boldsymbol{\tilde{c}}$ is the mistaken context. To achieve social adversary, we formulate the adversarial training between ego-driving policy and upper-level policies of background agents as a zero-sum game:
\begin{align}
    \beta^{l,*}_{1} &= \max_{\beta^l_{1}} \mathbb{E}_{\tau \sim \beta^l_{1}, \beta^u_i} [\sum_{t=0}^{T}\gamma^{t} R_{1}]  \label{equation:social-adversary-ego}   \\
    \beta^{u,*}_{i} &= \max_{\beta^u_{i}} \mathbb{E}_{\tau \sim \beta^l_{1}, \beta^u_i} [\sum_{t=0}^{T}\gamma^{t} (-R_{1})]
    \label{equation:social-adversary-adv}
\end{align}
For each background agent, its upper-level policy produces $\boldsymbol{\tilde{c}}$ to degrade the performance of ego agent. To maintain the efficiency among background agents, the adversarial policy only produces mistaken SVO of ego agent, while passing genuine SVOs of other background agents to its socially-aware policy. Therefore, the output of each adversarial policy is $\boldsymbol{\tilde{c}} = \left< \tilde{c}_1, \bar{c}_2, \dots, \bar{c}_n \right> $ and $\beta^u_i(\boldsymbol{\tilde{c}} | o_i) = \beta^u_i(\tilde{c}_1 | o_i) \prod_{j \in \mathcal{I}_n  \backslash \{1\}} \delta (c_j - \bar{c}_j)$.

After adversarial training with safety-critical data, the learned ego-policy $\beta^{l,*}_{1}$ is expected to be robust, enabling zero-shot transfer to unseen traffic flows. In addition, the learned background policies $\beta^{u,*}_{i}, \beta^{l,*}_{i}$ constitute the socially adversarial traffic flow.

\textbf{Network Architecture:}
The network architecture of ego policy conforms to that of the lower-level policy in Stage 1, as described in \Secref{section-method-flow:nn}. Adversarial communication policy also follows the same structure with the exception that it does not take SVOs as input and the length of dynamic vectors $V$ is now $5$.

\subsection{Training}

We apply Robust Adversarial Reinforcement Learning (RARL) \cite{pinto2017robust} to solve \Eqref{equation:social-adversary-ego} and \Eqref{equation:social-adversary-adv}.
RARL alternatively optimizes two policies using any single-agent algorithms and we utilize SAC to train each policy. Through this way, driving policy $\beta^{l}_{1}$ is learned and adversary policy $\beta^{u}_i$ is held constant in the first phase. In the second phase, we learn $\beta^{u}_i$ while $\beta^{l}_{1}$ is fixed.
For each phase, we collect data by rolling out $\beta^{l}_{1}$ and $\beta^{u}_i$ for several episodes and optimize one policy.

\textbf{Implementation Details:} 
We apply parameter sharing where all adversarial policies share the same network, i.e., $\beta^u_i(\tilde{c}_1 | o_i) = \beta^u(\tilde{c}_1 | o_i)$.
Noticeably, to apply adversary towards ego agent, the adversarial policy must be able to identify the ego agent in its observation $o_i$. However, agent $i$ cannot distinguish ego agent from its own perspective. To fix this, we transform all vectors in $o_i$ into the coordinate system of ego agent and apply adversary from the perspective of ego agent. Through this way, adversarial policy is conditioned on $o_1$. Similarly, we clip ego-observation $o_1$. Accordingly, background agents that are not at a certain distance relative to the ego agent are considered to have no impact on the ego agent, and their upper-level policy is the communication policy that produces genuine SVOs.

\subsection{Discussion on Existing Adversary Strategy under CPOSG}

\textbf{Existing Adversary Strategy:} In prior works, the background agent policy $\beta_i (a_i | o_i)$ attacks ego agent by the game:
\begin{align}
    & \max_{\beta_1} \mathbb{E}_{\tau \sim \beta_{1}, \beta_{i}} [\sum_{t=0}^{T}\gamma^{t} R_{1}]    \label{equation:minimax-general-sum-ego}  \\
    & \max_{\beta_i} \mathbb{E}_{\tau \sim \beta_{1}, \beta_{i}} [\sum_{t=0}^{T}\gamma^{t}( -R_{1} + \alpha_i R_i)]    \label{equation:minimax-general-sum-adv}
\end{align}
where $\alpha_i \geq 0$ is a hyperparameter. $R_i$ encourages background agent $i$ to behave normally including driving fast and avoiding failures except for collision. Formally, $R_i = \omega_1 R_{i,speed} + \omega_2 R_{i,fail} + \omega_2 \mathbf{1}( \text{Collision}(i) )$.

\textbf{Reformulation under CPOSG:} To better compare our adversarial traffic flow and existing works in adversarial attack, we rewrite the objective of background agents in \Eqref{equation:minimax-general-sum-adv} into an SVO style:
\begin{equation}
    \max_{\beta_i} \mathbb{E}_{\tau \sim \beta_{1}, \beta_{i}} [\sum_{t=0}^{T}\gamma^{t}(\cos c_i \cdot R_i + \sin c_i \cdot R_{1})]    \label{equation:min-general-sum-svo}
\end{equation}
where $c_i = -\arctan(1 / \alpha_i) \in [-90^\circ, 0^\circ)$. From this viewpoint, agents in their works can also be formulated as a CPOSG with agent-wise context $c_i$. Different from that in \Secref{section-method-flow:comp-reward}, in \Eqref{equation:min-general-sum-svo} each background agent only composes the reward of ego agent. Background policy adopts a hierarchical structure in which upper-level policy is communication policy that communicates a constant context. As such, the policy is directly proportional to its lower-level policy. Therefore, \Eqref{equation:minimax-general-sum-ego} and \Eqref{equation:minimax-general-sum-adv} are transformed into:
\begin{align}
    & \max_{\beta^l_1} \mathbb{E}_{\tau \sim \beta^l_{1}, \beta^l_{i}} [\sum_{t=0}^{T}\gamma^{t} R_{1}]   \label{equation:minimax-general-sum-svo-ego} \\
    & \max_{\beta^l_i} \mathbb{E}_{\tau \sim \beta^l_{1}, \beta^l_{i}} [\sum_{t=0}^{T}\gamma^{t}(\cos c_i \cdot R_i + \sin c_i \cdot R_{1})]
    \label{equation:minimax-general-sum-svo-adv}
\end{align}

\textbf{Comparative Discussion:} We highlight the key differences between our socially adversarial traffic flow and prior adversarial agents:
\begin{itemize}
    \item Interaction. Our approach assigns variable contexts, resulting in greater diversity in the traffic flow. Additionally, we do not train lower-level policies to be adversarial. Instead, we stimulate coordination among them, producing a highly efficient flow.
    \item Safety-criticism. Both our method and prior approaches utilize adversarial training between ego-driving policy and background policies. However, in our traffic flow, adversary stems from upper-level background policies, rather than lower-level ones. Furthermore, our traffic flow can avoid severe attack behaviors and maintains sufficient interaction by freezing lower-level policies of flow.
    \item Density. By balancing interactive and safety-critical behaviors, our method is capable of scaling up to dense traffic flows comprising dozens of background agents, unlike prior approaches that involve a smaller group of background agents.
\end{itemize}

%% file: sections/4-results.tex
\section{Experiments}  \label{section-exp}

\begin{figure*}[ht]
    \begin{center}
    \includegraphics[width=1.0\textwidth]{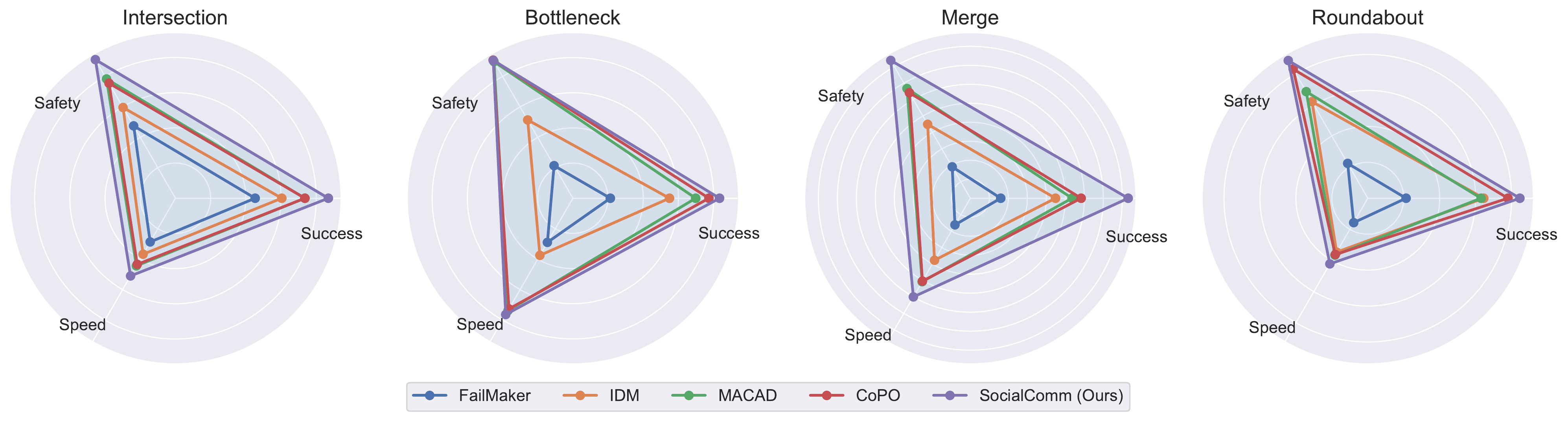}
    \end{center}
\vspace{-0.3cm}
\caption{Performance of various traffic flows. The radar graphs demonstrate three essential metrics including Success, Safety, and Speed. Each data point is computed by averaging $2000$ test episodes. The comprehensive performance of each traffic flow is determined by the area enclosed by these metrics.}
\label{figure:traffic-flow-polar}
\end{figure*}

In this section, we verify the proposed algorithm and pursue to answer the following three seminal questions:
\begin{itemize}
    \item Q1: Does the proposed socially-aware traffic flow enhance efficiency, preserve diversity, and elicit social behaviors (in \Secref{section-exp:q1})?
    \item Q2: Does the proposed socially adversarial traffic flow avoid over-strong adversaries and exhibit socially adversarial behaviors towards driving policy (in \Secref{section-exp:q2})?
    \item Q3: Can driving policy trained in our socially adversarial traffic flow improve the performance of zero-shot transfer (in \Secref{section-exp:q3})?
\end{itemize}

Before discussing these questions, we first describe some details including our environmental setup, definition of metrics, and comparative methods.

\subsection{Environmental Setup}  \label{section-exp:sim}

\textbf{Simulator}. We evaluate various methods using our driving simulator with four challenging scenarios: \texttt{intersection}, \texttt{bottleneck}, \texttt{merge}, and \texttt{roundabout}, as depicted in \Figref{figure:traffic_flow_behavior}.
Our driving simulator is 2D and aims to investigate single- and multi-agent driving behaviors, especially in dense traffic flows. Inspired by the trajectory prediction community \cite{liang2020learning, zhao2021tnt, gu2021densetnt}, our simulator utilizes sparse vectorized representation to capture the structural information of high-definition maps and agents. Compared to rasterized encoding which rasterizes the HD map elements together with agents into an image, vectorized representation is computation- and memory-efficient \cite{gao2020vectornet}.

\textbf{Termination Condition:} For an RL-oriented simulator, the termination condition of the environment needs delicate design. For single-agent environments, each episode ends when ego agent is terminated. For multi-agent environments, each episode ends when all agents are terminated. Once an agent is terminated, it is removed from the scenario. Specifically, one agent is terminated if it reaches its destination, encounters catastrophic failures, or survives until timeout. Catastrophic failures include collision with other agents, deviation from drivable area, driving too far from global path, and crashing into wrong lane. The maximum steps for one episode are $T$ and when an agent survives $T$ steps in the environment, we call it ``timeout". An agent is marked as a success only when it passes through the interaction zone \footnote{We name each scenario with its interaction zone. For instance, the interaction zone in $\texttt{bottleneck}$ is the bottleneck. Visualization of interaction zones can be found in \url{https://github.com/alibaba-damo-academy/universe/blob/main/README.md}.} before the timeout occurs.

\begin{table}[t]
\renewcommand{\arraystretch}{1.1}
\centering
\caption{Hyperparameters of SAC}
\label{table:sac-param}
\vspace{1mm}
\begin{tabular}{l l }
\toprule
{Parameter} & Value\\
\midrule
    optimizer &Adam\\
    actor learning rate & $1 \cdot 10^{-4}$\\
    critic learning rate & $5 \cdot 10^{-4}$\\
    tune learning rate & $1 \cdot 10^{-4}$\\
    discount ($\gamma$) &  $0.9$\\
    batch size & $128$\\
    replay buffer size & $10^6$\\
    nonlinearity & ReLU\\
    target smoothing coefficient & $0.005$\\
    target update interval & $1$\\
    gradient steps & $1$\\
\bottomrule
\end{tabular}
\end{table}

\textbf{Training Framework:} We use PyTorch 1.9.1 \footnote{Official documentation is at \url{https://pytorch.org/docs/1.9.1}.} to implement policy architectures and use Adam \cite{kingma2015adam} to optimize policies. All models are trained with one Nvidia RTX 3090. Detailed parameters of Soft-Actor-Critic (SAC) are shown in \Tabref{table:sac-param}. We use precisely the same parameters for both single- and multi-agent policy learning.
We use Ray \cite{moritz2018ray} to implement parallel sampling and use 16 workers with different seeds to sample experiences in parallel across four scenarios. Each scenario is assigned 4 workers.

\subsection{Metrics}

\begin{figure*}[ht]
    \begin{center}
    \includegraphics[width=1.0\textwidth]{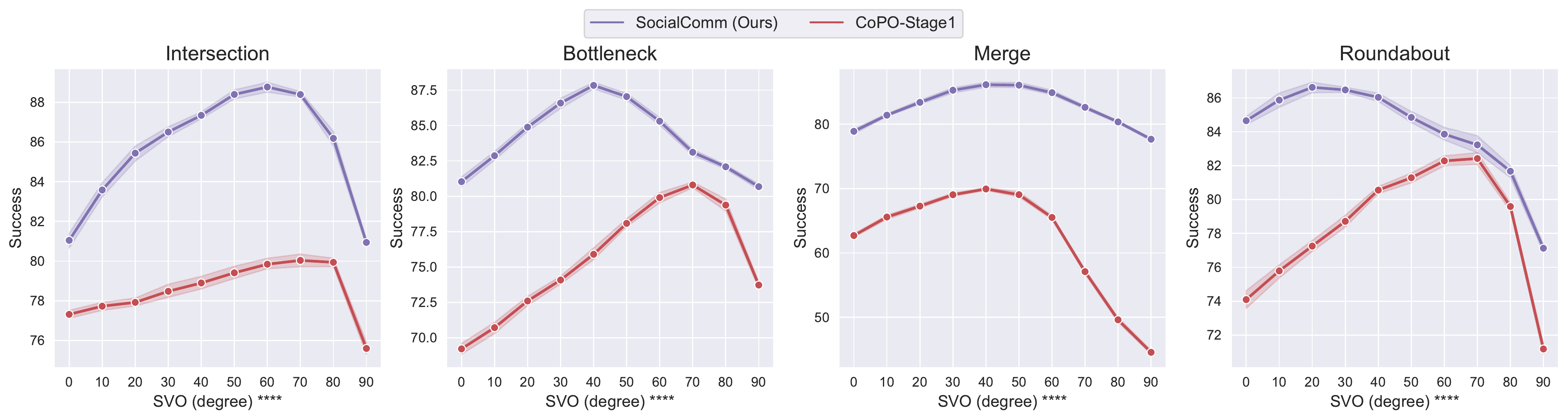}

    \end{center}
\vspace{-0.3cm}
\caption{Success rates of CoPO and SocialComm (Ours). The figure reports success rates with $95\%$ confidence interval w.r.t. SVOs. Each data point is computed by averaging $2000$ test episodes. A higher value indicates better performance. Statistical significance is assessed using the Paired T-Test where $\star$$\star$$\star$$\star$ means P-Value $p < 0.0001$.}
\label{figure:traffic-flow-svo}
\end{figure*}

\begin{figure}[t]
    \begin{center}
    \hspace{-0.3cm}\includegraphics[width=0.5\textwidth]{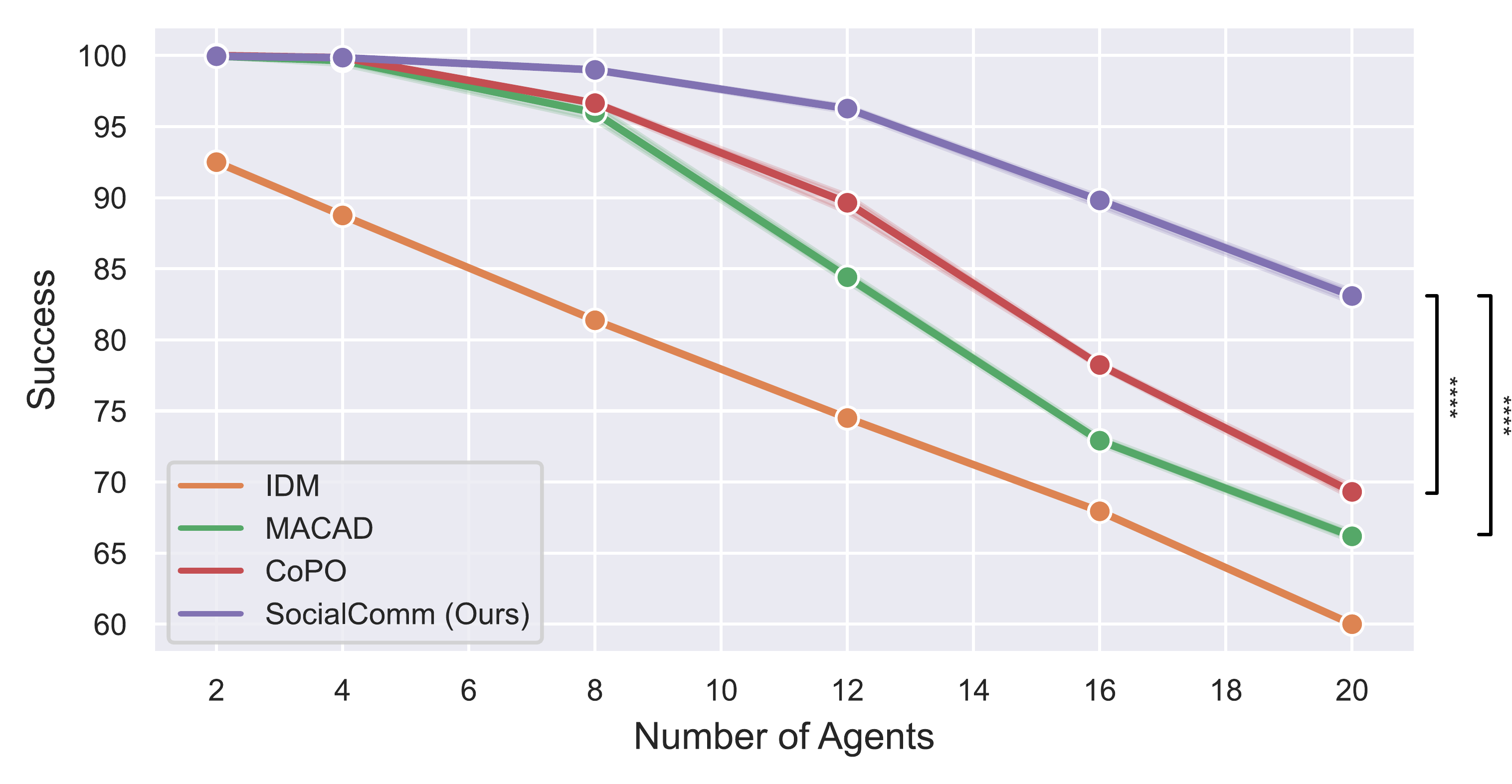}
    \end{center}
\vspace{-0.3cm}
\caption{Success rates of various traffic flows in \texttt{merge}. The figure reports success rates with $95\%$ confidence interval w.r.t. the number of agents. Each data point is computed by averaging $2000$ test episodes. A higher value indicates better performance. Statistical significance is assessed using the Paired T-Test where $\star$$\star$$\star$$\star$ means P-Value $p < 0.0001$.}
\label{figure:flow-vary-number-merge}
\end{figure}

We consider three kinds of widely-accepted and general metrics which are closely related to termination conditions and are represented as percentages:

\begin{itemize}
    \item \textbf{\emph{Success}} \cite{dosovitskiy2017carla, rhinehart2019deep, chen2019attention, cai2019lets, wang2020learning, chen2020learning, wu2021flow, wang2021imitation, peng2021learning, wang2022domain}: For ego-driving policy evaluation, this is the success rate of ego agent. For traffic flow evaluation, this is the success rate of the whole flow which is averaged by the success rate of each agent.
    \item \textbf{\emph{Safety}} \cite{peng2021learning}: The complement of Catastrophic Failures including \textbf{\emph{Collision}} \cite{chen2019attention, cai2019lets, wang2020learning, chen2020learning, suo2021trafficsim}, \textbf{\emph{Off Road}} \cite{rhinehart2019deep, chen2019attention, wang2020learning}, \textbf{\emph{Off Route}} \cite{chen2019model, toromanoff2020end}, and \textbf{\emph{Wrong Lane}} \cite{rhinehart2019deep, wang2020learning}. Collision means collision between agents. Off Road means deviation from drivable area. Off Route means driving too far from global path. Wrong Lane means crashing into wrong lane. For ego-driving policy evaluation, this is the safety of ego agent. For traffic flow evaluation, this is the safety of the whole flow which is averaged by the safety of each agent. Note that Safety is not always equal to Success, since there exists a situation the agent does not encounter catastrophic failures but stays in the interaction zone at the end of one episode. For this agent, it is safe but does not succeed.
    \item \textbf{\emph{Speed}} \cite{chen2019attention, cai2019lets, moghadam2020end, wu2021flow}: For ego-driving policy evaluation, this is the average speed of ego agent among each episode. For traffic flow evaluation, this is the average speed of the whole flow which is averaged by the average speed of each agent. 
\end{itemize}

\vspace{-0.1cm}

\subsection{Benchmark Methods}

\begin{figure*}[t]
\begin{center}
\includegraphics[width=0.95\textwidth]{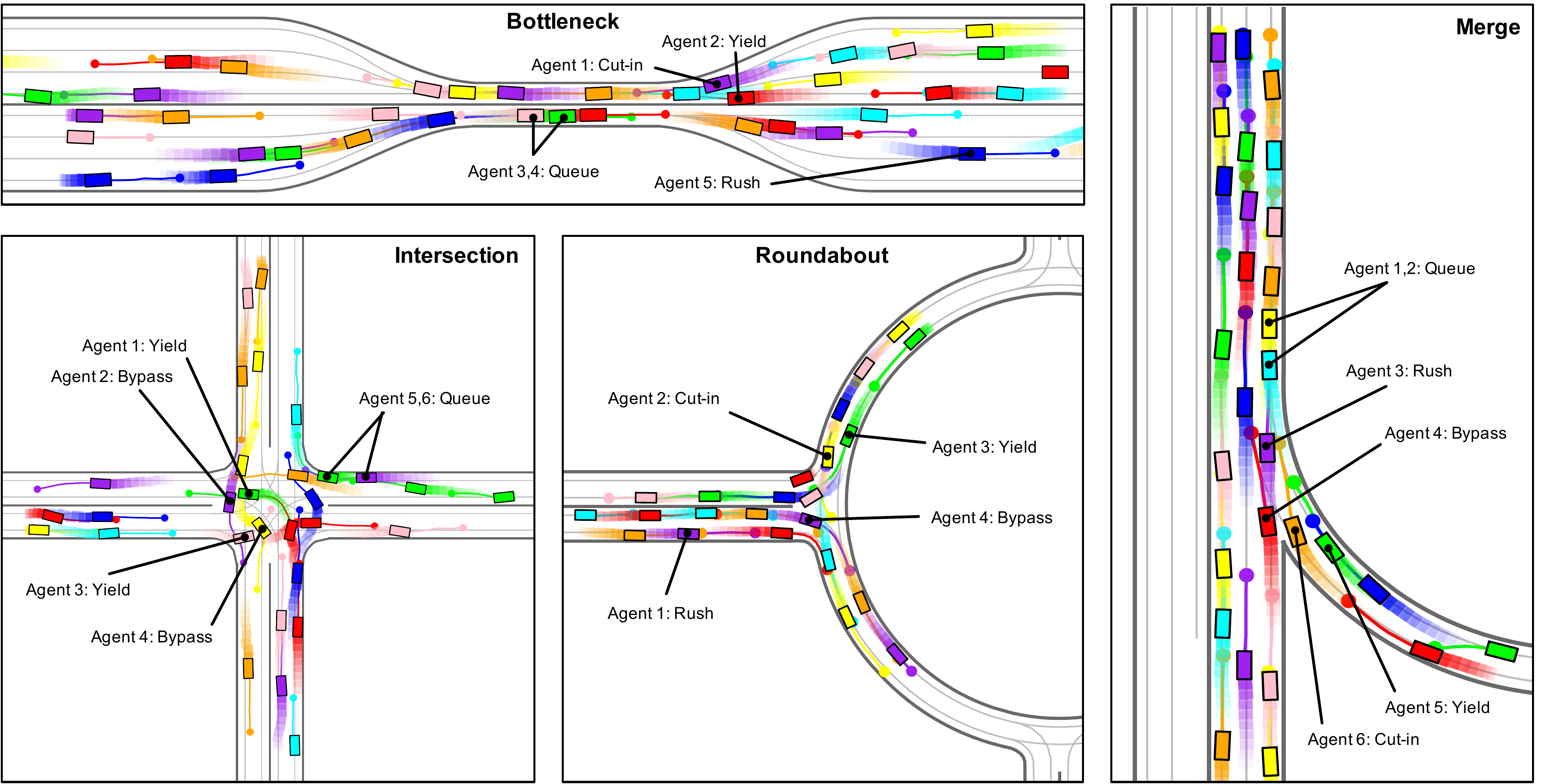}
\end{center}
\vspace{-0.3cm}
\caption{Social behaviors of our SocialComm. The figure demonstrates behaviors at a given time step. Colored boxes with shaded tails represent historical trajectories and curves with dots represent future trajectories.}
\label{figure:traffic_flow_behavior}
\end{figure*}

For comparison of traffic flow, we implement six representative traffic flows. Traffic flows highlighted in {\color{DeepSkyBlue} blue are non-adversarial} and those highlighted in {\color{red} red are adversarial}, as following:

\begin{itemize}
    \item \textbf{{\color{DeepSkyBlue} \emph{Intelligent Driver Model (IDM)} \cite{treiber2013traffic}}}: Each agent strictly follows its global path and is controlled by an analytical controller which uses one single differential equation to model longitudinal movements.
    \item \textbf{{\color{DeepSkyBlue} \emph{MACAD} \cite{palanisamy2020multi}}}: This method is modeled as POSG and each agent aims to maximize its individual reward. The number of agents in their origin paper is 3. For a fair comparison, we implement this method with the same number of agents as other flows.
    \item \textbf{{\color{DeepSkyBlue} \emph{CoPO} \cite{peng2021learning}}}: This traffic flow can be modeled as CPOSG and incorporates SVO into MARL. A two-stage framework is proposed. In CoPO-Stage1, each agent has a distinct SVO sampled from uniform distribution and is driven by a hierarchical policy, where upper-level communication policy is self-visible and lower-level policy is conditioned on its own SVO. In CoPO-Stage2, the hierarchical policy is frozen. All agents share a common Gaussian distribution which is learned to enhance the sum of each agent's individual reward, serving as a central coordinator.
    \item \textbf{{\color{DeepSkyBlue} \emph{SocialComm (Ours)}}}: Socially-aware traffic flow proposed in \Secref{section-method-flow}.
    \item \textbf{{\color{red} \emph{FailMaker} \cite{wachi2019failure}}}: This adversary strategy can be modeled by CPOSG with objective \Eqref{equation:min-general-sum-svo}. The number of agents in their origin paper is 1 or 2. For a fair comparison, we implement this method with the same number of agents as other flows. Furthermore, to avoid excessive attack behaviors and increase diversity, we assign a distinct SVO for each agent from a uniform distribution ranging from $-90^\circ$ to $90^\circ$.
    \item \textbf{{\color{red} \emph{SocialCommAdv (Ours)}}}: Socially adversarial traffic flow proposed in \Secref{section-method-adv}.
\end{itemize}

For comparison of ego-driving policy, we take two training pipelines including {\color{DeepSkyBlue} reinforcement learning (RL)} and {\color{red} robust adversarial reinforcement learning (RARL)} to train ego-driving policies under various traffic flows. {\color{DeepSkyBlue} RL} can only be applied in non-adversarial traffic flows including {\color{DeepSkyBlue} IDM}, {\color{DeepSkyBlue} MACAD}, {\color{DeepSkyBlue} CoPO}, and {\color{DeepSkyBlue} SocialComm}. {\color{red} RARL} can only be applied in {\color{red} FailMaker} and {\color{red} SocialCommAdv}. Each trained ego-driving policy is identified by its corresponding training pipeline and traffic flow, such as {\color{DeepSkyBlue} RL/IDM} and {\color{red} RARL/SocialCommAdv}.

To ensure a fair comparison of various methods, we generate $200$ cases offline with different global paths, initial poses, and SVOs for each scenario. To obtain reliable results, we run each method on these cases $10$ times with different seeds, in total $2000$ test episodes.

\subsection{Q1: Performance of Traffic Flows}  \label{section-exp:q1}

In this part, we validate the interaction level of various traffic flows from two aspects: efficiency and diversity, and demonstrate emerged social behaviors in our SocialComm.

\textbf{Efficiency:} \Figref{figure:traffic-flow-polar} presents metrics of various traffic flows in all scenarios from three dimensions: Success, Safety, and Speed. Each metric is normalized to its maximal and minimal values in each scenario, wherein a value further from the center indicates better performance. Data in each metric is computed by the average of $2000$ test episodes, indicating the statistical significance of comparison.
Comprehensively, the proposed SocialComm achieves the highest performance in all scenarios. In comparison, FailMaker achieves the lowest performance and underperforms other traffic flows by a large margin in almost all scenarios. These results indicate that fully-visible communication policy can indeed improve overall performance and overlooking interaction in the traffic flow leads to poor success rates.

\textbf{Density:} Note that in \texttt{merge}, our traffic flow outperforms other methods by a large margin as shown in \Figref{figure:traffic-flow-polar}. Success and Safety are almost equally improved. The reason is that the poses of all agents in \texttt{merge} are much closer compared to other scenarios, making it hard for comparative traffic flows to maintain a high level of safety. We further illustrate this phenomenon by \Figref{figure:flow-vary-number-merge}, wherein we evaluate success rates of traffic flows with different densities in \texttt{merge}. FailMaker is omitted due to its relatively low success rate.
For each traffic flow, we evaluate $6$ different densities and apply Paired T-Test with $12000$ samples.
One can find that comparative traffic flows dramatically decrease as the number of agents increases. On the contrary, SocialComm is less sensitive to the variation of traffic flow densities. This again confirms that our traffic flow is statistically significantly more efficient than other flows, especially in denser settings.

\textbf{Diversity:} Under CPOSG, the diversity of traffic flow stems from the distribution of SVOs. Accordingly, we compare CoPO-Stage1 and SocialComm which utilizes a uniform distribution of SVOs. Results are shown in \Figref{figure:traffic-flow-svo}. For ease of analysis, we let all agents in traffic flows share the same SVO and sample it at regular intervals ranging from $0^\circ$ to $90^\circ$. For each traffic flow, $10$ different SVOs are evaluated, resulting in $20000$ samples for Paired T-Test.
As observed in the results, SocialComm outperforms CoPO-Stage1 for all SVOs in all scenarios. Moreover, most data points in SocialComm exhibit higher success rates compared to the maximum data points in CoPO-Stage1. This further validates that increasing the visibility of communication policy improves success rate of the whole flow. Furthermore, in most scenarios, success rates in SocialComm show less variation w.r.t. SVOs compared to CoPO-Stage1. This implies that SocialComm holds high efficiency while preserving diversity in a statistically significant manner. Therefore, our method can eliminate the trade-off between these two factors that are often present in previous traffic flows, making our dense traffic flow more interactive.

\textbf{Social Behaviors:} \Figref{figure:traffic_flow_behavior} demonstrates emerged social behaviors in our SocialComm.
As one can see, agents in SocialComm exhibit diverse social behaviors such as queueing at the narrow crossing, rushing at open areas, and yielding to avoid crashes.
In \texttt{intersection}, agent 1 and 3 yield to agent and 4, respectively, to allow them to bypass and drive faster. In \texttt{bottleneck}, agent 3 and 4 queue to pass through the bottleneck. In \texttt{merge}, agent 1 and 2 are queueing along the traffic flow. Agent 6 maneuvers a cut-in which agent 5 yields to. In \texttt{roundabout}, agent 1 approaches the roundabout rapidly and agent 2 cuts in front of agent 3, causing it to yield.

\subsection{Q2: Effect of Social Adversary}  \label{section-exp:q2}

\begin{figure*}[htp]
    \begin{center}
    \includegraphics[width=0.95\textwidth]{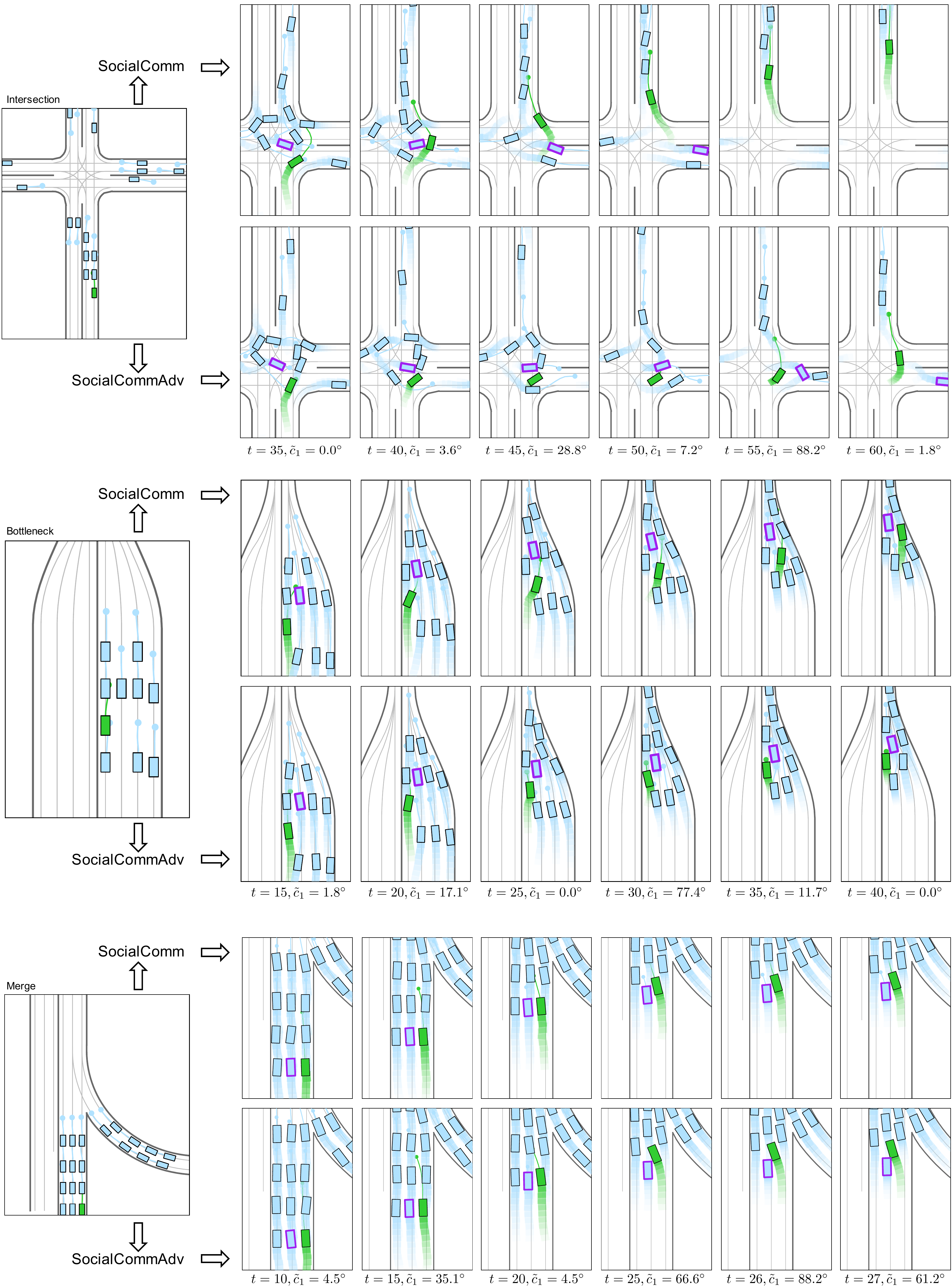}
    \end{center}
\vspace{-0.3cm}
\caption{Socially adversarial behaviors towards driving agent generated by our SocialCommAdv. In each frame shot, ego agent is marked as {\color{LimeGreen} green}. Background agents are marked as {\color{DeepSkyBlue} blue}. Agent with {\color{Purple} purple} box exhibits substantial adversarial behaviors towards ego agent.}
\label{figure:social-adversary-behavior}
\end{figure*}

\begin{table}[t]
\renewcommand\arraystretch{1.2}
\centering
\caption{Effects of adversaries on Success$^{*}$}
\label{table:adversary-compare-success}
\begin{threeparttable}
\begin{tabular}{ccccc}
    \toprule[1pt]
    \textbf{Method} & \textbf{Scenario} & \textbf{{\color{DeepSkyBlue} SC}} & \textbf{{\color{red} SCA}} & \textbf{{\color{red} FM}} \\
    \hline
    \multirow{4}{*}{RL/IDM}
    & \texttt{I}   &  76.5  &  74.8 (-1.7)  &  55.2 (-21.3) \\
    & \texttt{B}   &  52.0  &  50.8 (-1.2)  &  20.6 (-31.4) \\
    & \texttt{M}   &  64.5  &  62.8 (-1.7)  &  9.2 (-55.3) \\
    & \texttt{R}   &  75.0  &  72.8 (-2.2)  &  19.6 (-55.4) \\

    \hline
    \multirow{4}{*}{RL/MACAD}
    & \texttt{I}   &  83.6  &  82.6 (-1.0)  &  52.1 (-31.5) \\
    & \texttt{B}   &  75.2  &  73.6 (-1.6)  &  4.6 (-70.6) \\
    & \texttt{M}   &  71.2  &  69.6 (-1.6)  &  9.1 (-62.1) \\
    & \texttt{R}   &  73.6  &  71.6 (-2.0)  &  29.6 (-44.0) \\

    \hline
    \multirow{4}{*}{RL/CoPO}
    & \texttt{I}   &  81.6  &  78.8 (-2.8)  &  51.0 (-30.6) \\
    & \texttt{B}   &  71.6  &  69.8 (-1.8)  &  6.5 (-65.1) \\
    & \texttt{M}   &  70.1  &  66.2 (-3.9)  &  10.0 (-60.1) \\
    & \texttt{R}   &  82.1  &  80.8 (-1.3)  &  21.0 (-61.1) \\

    \hline
    \multirow{4}{*}{RL/SocialComm}
    & \texttt{I}   &  87.2  &  85.2 (-2.0)  &  51.4 (-35.8) \\
    & \texttt{B}   &  91.2  &  85.2 (-6.0)  &  2.4 (-88.8) \\
    & \texttt{M}   &  84.8  &  80.2 (-4.6)  &  8.4 (-76.4) \\
    & \texttt{R}   &  87.8  &  84.7 (-3.1)  &  17.0 (-70.8) \\

    \bottomrule[1pt]
\end{tabular}

\begin{tablenotes}
    \footnotesize
    \item[*] Each data point is computed by averaging $2000$ test episodes. \texttt{I}, \texttt{B}, \texttt{M}, and \texttt{R} are abbreviations of each scenario. Results not in parentheses are performances in {\color{DeepSkyBlue} SocialComm}, {\color{red} SocialCommAdv}, and {\color{red} FailMaker}, abbreviated as {\color{DeepSkyBlue} SC}, {\color{red} SCA}, and {\color{red} FM}, respectively. Results in parentheses indicate performance degradations under adversaries.
\end{tablenotes}
\end{threeparttable}

\end{table}

In this part, we demonstrate the performance degradation of different ego-driving policies under adversaries and exhibit socially adversarial behaviors towards ego agent produced by the proposed SocialCommAdv.

\textbf{Performance Degradation under Adversary:}
\Tabref{table:adversary-compare-success} and \Tabref{table:adversary-compare-speed} illustrate the impact of adversaries on ego-driving policies. Ego-driving policies are trained in non-adversarial traffic flows (including {\color{DeepSkyBlue} IDM}, { \color{DeepSkyBlue}MACAD}, {\color{DeepSkyBlue} CoPO}, and {\color{DeepSkyBlue} SocialComm}) and are evaluated in {\color{DeepSkyBlue} SocialComm}, {\color{red} SocialCommAdv}, and {\color{red} FailMaker}.
Data in parentheses denote the performance degradation in two adversarial traffic flows ({\color{red} SocialCommAdv} and {\color{red} FailMaker}) compared to {\color{DeepSkyBlue} SocialComm}. Note that ego-driving policies trained in {\color{red} SocialCommAdv} and {\color{red} FailMaker} are not included in this table since policies trained under adversaries can resist disturbances and may obtain higher performances in adversarial flows, which could skew the investigation of the effect of adversaries.
The results indicate that success rates and speeds of all ego-driving policies decrease under adversaries in all scenarios, thus validating the effectiveness of social adversary. Moreover, the effect of {\color{red} FailMaker} is an order of magnitude higher than that of {\color{red} SocialCommAdv}, implying that prior adversary strategy results in strong disturbances towards ego agent. In contrast, {\color{red} SocialCommAdv} produces moderate adversarial effects that avoid excessively strong attacks and maintains sufficient interaction, thanks to the freezing of the lower-level policy in our flow. Although the performance degradation of ego-driving policies in {\color{red} SocialCommAdv} is moderate, we find that it plays a critical role in improving the performance of zero-shot transfer in \Secref{section-exp:q3}.

\textbf{Socially Adversarial Behaviors:} \Figref{figure:social-adversary-behavior} displays several qualitative results on how {SocialCommAdv} impedes ego agent to finish its own task. In each scenario, figures in the first row depict behaviors under {SocialComm}, and figures in the second row present behaviors under {SocialCommAdv}. Ego-driving policies are trained in {SocialComm}. Leftmost figures demonstrate initial poses of all agents, which are identical in both {SocialComm} and {SocialCommAdv}.
$t$ represents the time step and $\tilde{c}_1$ denotes the mistaken SVO of ego agent produced by adversarial communication policy.
In \texttt{intersection}, at $t = 40$, background agents ahead of ego agent group together to impede ego agent from bypassing the traffic flow. At $t = 50$, the impact agent executes a cut-in, compelling the ego agent to stop.
In \texttt{bottleneck}, under {SocialComm}, ego agent performs a lane change to pass the bottleneck more efficiently at $t = 25, 30, 35$. While in {SocialCommAdv}, the impact agent slows down at $t = 20$ and occupies the lane that ego agent aims to change to at $t = 25, 30, 35$. Consequently, the ego agent stays at its original lane and passes the bottleneck behind other agents. In \texttt{merge}, there exists a gap between impact agent and other background agents. To improve efficiency, ego agent identifies this gap and executes a cut-in, thereby making the impact vehicle yield. In {SocialCommAdv}, impact agent collides with ego agent in a natural way, leading to the failure of ego agent.

\begin{table}[t]
\renewcommand\arraystretch{1.2}
\centering
\caption{Effects of adversaries on Speed$^{*}$}
\label{table:adversary-compare-speed}
\begin{threeparttable}
\begin{tabular}{ccccc}
    \toprule[1pt]
    \textbf{Method} & \textbf{Scenario} & \textbf{{\color{DeepSkyBlue} SC}} & \textbf{{\color{red} SCA}} & \textbf{{\color{red} FM}} \\
    \hline
    \multirow{4}{*}{RL/IDM}
    & \texttt{I}   &  47.4  &  46.4 (-1.0)  &  37.3 (-10.1) \\
    & \texttt{B}   &  58.6  &  57.5 (-1.1)  &  34.5 (-24.1) \\
    & \texttt{M}   &  51.6  &  51.2 (-0.4)  &  15.1 (-36.5) \\
    & \texttt{R}   &  38.3  &  38.3 (0.0)  &  20.2 (-18.1) \\
    
    \hline
    \multirow{4}{*}{RL/MACAD}
    & \texttt{I}   &  49.8  &  49.7 (-0.1)  &  36.1 (-13.7) \\
    & \texttt{B}   &  74.3  &  73.9 (-0.4)  &  24.3 (-50.0) \\
    & \texttt{M}   &  54.7  &  54.2 (-0.5)  &  17.6 (-37.1) \\
    & \texttt{R}   &  39.4  &  39.6 (0.2)  &  25.1 (-14.3) \\
    
    \hline
    \multirow{4}{*}{RL/CoPO}
    & \texttt{I}   &  49.3  &  47.6 (-1.7)  &  34.9 (-14.4) \\
    & \texttt{B}   &  71.4  &  70.0 (-1.4)  &  27.7 (-43.7) \\
    & \texttt{M}   &  55.6  &  53.5 (-2.1)  &  19.4 (-36.2) \\
    & \texttt{R}   &  42.9  &  42.3 (-0.6)  &  22.9 (-20.0) \\
    
    \hline
    \multirow{4}{*}{RL/SocialComm}
    & \texttt{I}   &  52.1  &  51.0 (-1.1)  &  36.5 (-15.6) \\
    & \texttt{B}   &  81.5  &  78.5 (-3.0)  &  24.3 (-57.2) \\
    & \texttt{M}   &  60.9  &  59.4 (-1.5)  &  16.8 (-44.1) \\
    & \texttt{R}   &  43.4  &  42.5 (-0.9)  &  19.9 (-23.5) \\

    \bottomrule[1pt]
\end{tabular}

\begin{tablenotes}
    \footnotesize
    \item[*] Each data point is computed by averaging $2000$ test episodes. \texttt{I}, \texttt{B}, \texttt{M}, and \texttt{R} are abbreviations of each scenario. Results not in parentheses are performances in {\color{DeepSkyBlue} SocialComm}, {\color{red} SocialCommAdv}, and {\color{red} FailMaker}, abbreviated as {\color{DeepSkyBlue} SC}, {\color{red} SCA}, and {\color{red} FM}, respectively. Results in parentheses indicate performance degradations under adversaries.
\end{tablenotes}
\end{threeparttable}

\vspace{-0.3cm}
\end{table}

\begin{figure*}[htp]
	\centering
    \subfloat[Success]{
		\includegraphics[width=0.98\textwidth]{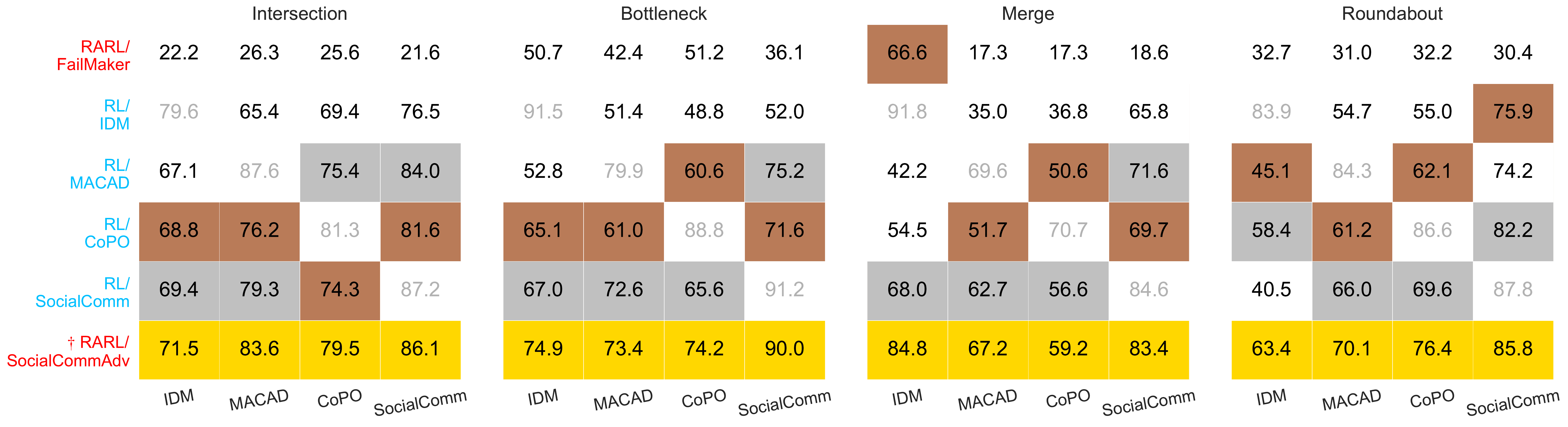}
    } \\

	\vspace{-0.3cm}
    \subfloat[Speed]{
		\includegraphics[width=0.98\textwidth]{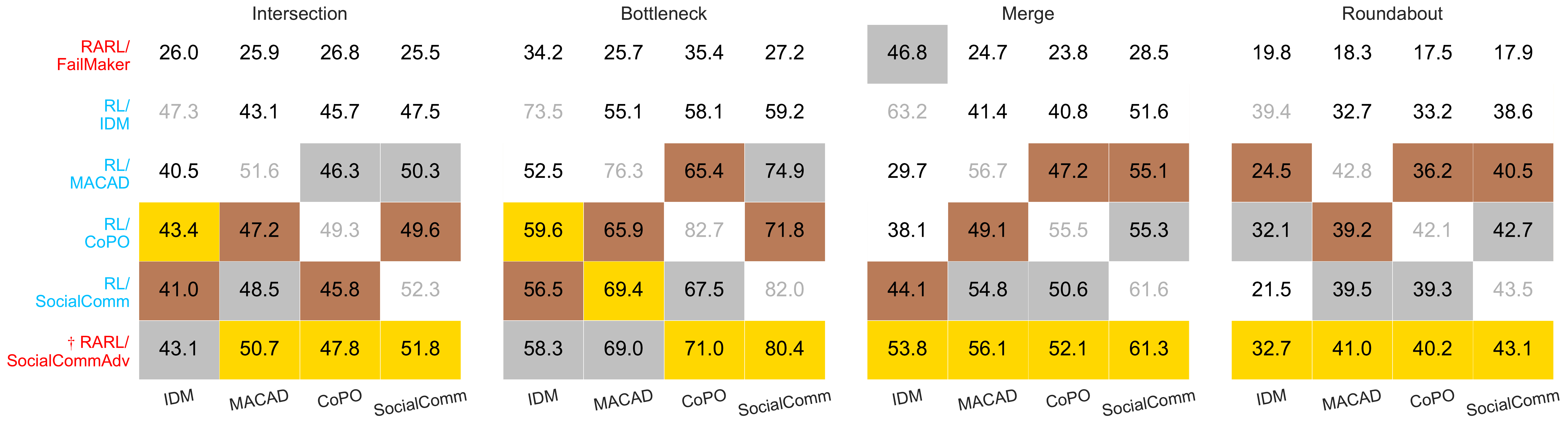}
    }

\caption{Zero-shot transfer performance of each ego-driving policy in each evaluating traffic flow.
    Heatmaps report success rates and speeds in various traffic flows for ego agents. Each data point is computed by averaging $2000$ test episodes. In each column, data filled with {\color{Gold} gold}, {\color{Silver} silver}, and {\color{Copper} copper} occupy the first, second, and third positions, respectively. Ego-driving policies trained with our social adversarial training are marked with a ``$^\dagger$". Data marked as grey indicates that training and evaluating traffic flows are the same and do not participate in ranking.
}
\label{figure:ood-performance}
\end{figure*}

\subsection{Q3: Zero-shot Transfer Performance of Driving Policies}   \label{section-exp:q3}

\textbf{Cross-Validation:} To evaluate zero-shot transfer in unseen traffic flows, we employ cross-validation. We train ego-driving policies in all traffic flows and evaluate each trained policy in four non-adversarial traffic flows - {IDM}, {MACAD}, {CoPO}, and {SocialComm}.
Success rates and Speeds of all evaluations are illustrated in \Figref{figure:ood-performance}.
To ensure statistical significance in the cell-to-cell comparison, we compute the data by averaging $2000$ test episodes for each cell.
Grey-marked data reflect cases where the evaluation was performed in the same training traffic flow. Remaining data reveal performances of zero-shot transfer. Note that in each heatmap, data marked as grey always achieve highest performances since an ego-driving policy tends to overfit to its training traffic flow and the performance degrades when transferred to unseen flows. Therefore, these results serve as upper bounds on performances of different ego-driving policies in each traffic flow and are excluded from ranking, concentrating on zero-shot transfer.

Overall, ego-driving policy trained in our SocialCommAdv achieves highest success rates in all scenarios and highest speeds in almost all scenarios. This indicates that our socially adversarial traffic flow has substantially enhanced the zero-shot transfer performance of ego-driving policy, contributing to the balance of efficiency, diversity, and adversary.

Regarding the effect on efficiency, one can observe the comparison among {RL/IDM}, {RL/MACAD}, {RL/CoPO}, and {RL/SocialComm}. Ego-driving policy located lower in each heatmap achieves higher performances in almost all evaluating traffic flows. This reveals that the zero-shot performances of these ego-driving policies are aligned with the efficiencies of their respective training traffic flows. This result suggests that improving the efficiency of the traffic flow can lead to better zero-shot transfer performance of ego-driving policy.

Regarding the effect on diversity, we compare the pair ({RL/IDM}, {RL/MACAD}) with ({RL/CoPO}, {RL/SocialComm}), where (CoPO, SocialComm) utilizes variable contexts, and (IDM, MACAD) utilizes constant context. As illustrated in our results, performances of ({RL/CoPO}, {RL/SocialComm}) are superior to those of ({RL/IDM}, {RL/MACAD}), indicating the effect of assigning variable contexts. Furthermore, {RL/SocialComm} outperforms {RL/CoPO}, implying that more diverse variable contexts lead to higher zero-shot transfer performance.

Regarding the effect on adversary, we first compare {RL/SocialComm} with {RARL/SocialCommAdv}, where {RARL/SocialCommAdv} outperforms {RL/SocialComm} in all scenarios. This result supports the effectiveness of social adversary on an interactive traffic flow.
We then compare {RARL/SocialCommAdv} with {RARL/FailMaker} and find a significant gap. RARL/FailMaker demonstrates worst zero-shot transfer performances. This is due to the lack of modeling interaction in FailMaker, which leads to over-strong attack behaviors and makes it fragile for {RARL/FailMaker} to unseen interactive behaviors.

\begin{figure*}[tp]
	\centering
    \subfloat[Statistical Significance of Success]{
		\includegraphics[width=0.98\textwidth]{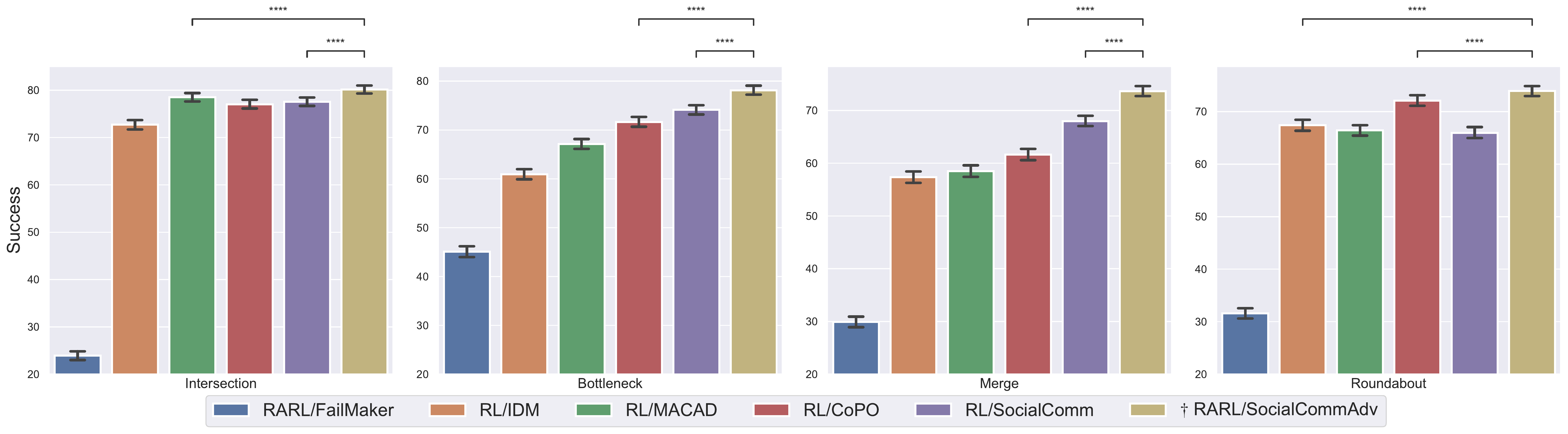}
    }    \\
	\vspace{-0.3cm}
    \subfloat[Statistical Significance of Speed]{
		\includegraphics[width=0.98\textwidth]{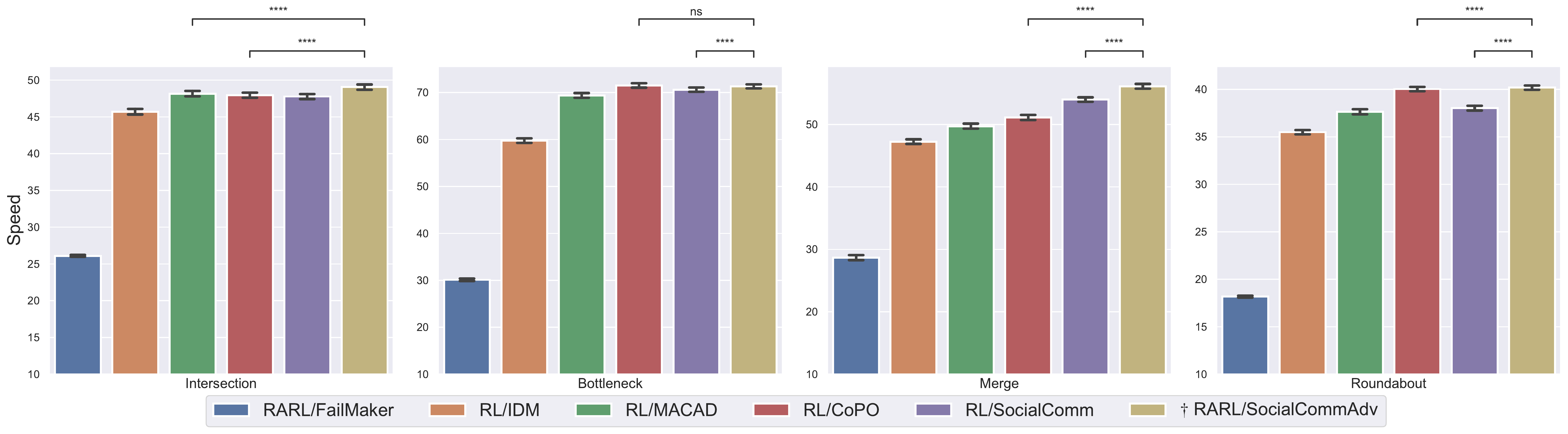}
    }

\caption{Zero-shot transfer performance of each ego-driving policy.
    Barcharts report success rates and speeds for ego agents with $95\%$ confidence interval. Each data point is computed by averaging $8000$ test episodes. A higher value indicates better performance. Statistical significance is assessed using the Paired T-Test where ns means $p \leq 1$, $\star$ means $p < 0.05$, $\star$$\star$ means $p < 0.01$, $\star$$\star$$\star$ means $p < 0.001$, $\star$$\star$$\star$$\star$ means $p < 0.0001$.
}
\label{figure:ood-performance-ss}
\end{figure*}

\textbf{Zero-shot Transfer of Ego-Driving Policy:} In \Figref{figure:ood-performance-ss}, we compare various ego-driving policies comprehensively. We group the evaluation results of $4$ non-adversarial traffic flows (IDM, MACAD, CoPO, and SocialComm) and perform Paired T-Test with two comparative methods that have the nearest means to our RARL/SocialCommAdv.
For each ego-driving policy, $8000$ samples are used to compute the mean and apply Paired T-Test.
Importantly, for the evaluation of each comparative method (except RARL/FailMaker), performance in its training, seen flow is taken into consideration, whereas our method is evaluated in completely unseen traffic flows. Remarkably, our method outperforms even the other approaches that are evaluated containing their training traffic flows. For a total of $16$ tests, $15$ tests are statistically significant with P-Value $p < 0.0001$. This reveals that our RARL/SocialCommAdv achieves superior zero-shot transfer performance statistically significantly.

%% file: sections/5-conclusion.tex
\section{Conclusion}  \label{section-conclusion}

This paper proposes a socially adversarial traffic flow that can improve the zero-shot transfer capability of driving policy. To achieve this, we first model the traffic flow as Contextual POSG and assign SVOs as contextual information.
We then two-stage framework. In Stage 1, we develop socially-aware traffic flow. Each agent in this traffic flow is driven by a hierarchical policy where communication policy produces genuine SVOs. We train lower-level policies via independent policy learning. In Stage 2, we develop socially adversarial traffic flow. The lower-level policies are learned lower-level policies in Stage 1. Communication policy for each agent produces mistaken SVOs. Ego agent interacts with this traffic flow. Ego-driving policy is jointly trained with communication policies of other agents adversarially.
We validate the performance of traffic flow comprehensively in four challenging scenarios: \texttt{intersection}, \texttt{bottleneck}, \texttt{merge}, and \texttt{roundabout}. Cross-validation verifies the superior performance of zero-shot transfer to unseen traffic flows of our method.

In the future, we plan to further improve the safety of our traffic flow by incorporating safe RL techniques \cite{brunke2022safe} to eliminate catastrophic failures of the traffic system and investigate the transfer capability of driving policy under this traffic flow.